\documentclass{article}

\usepackage{graphicx}
\usepackage{verbatim}
\usepackage{latexsym}
\usepackage{qtree}
\usepackage{algorithmic}
\usepackage{amssymb}

\begin{document}

\title{A Decomposition Approach to Multi-Vehicle Cooperative Control}

\author{Matthew Earl\thanks{Corresponding author. email: 
{\tt\small mge1@cornell.edu}}
~and Raffaello D'Andrea}

\date{}
\maketitle

\begin{abstract}
We present methods that generate cooperative strategies for
multi-vehicle control problems using a decomposition approach.  By
introducing a set of tasks to be completed by the team of vehicles and
a task execution method for each vehicle, we decomposed the problem
into a combinatorial component and a continuous component.  The
continuous component of the problem is captured by task execution, and
the combinatorial component is captured by task assignment.  In this
paper, we present a solver for task assignment that generates
near-optimal assignments quickly and can be used in real-time
applications.  To motivate our methods, we apply them to an
adversarial game between two teams of vehicles. One team is governed
by simple rules and the other by our algorithms. In our study of this
game we found phase transitions, showing that the task assignment
problem is most difficult to solve when the capabilities of the
adversaries are comparable.  Finally, we implement our algorithms in a
multi-level architecture with a variable replanning rate at each level
to provide feedback on a dynamically changing and uncertain
environment. 
\end{abstract}

\section{Introduction}

Using a team of vehicles to accomplish an objective can be effective
for problems involving a set of tasks distributed in space and time.
Examples of such problems include multi-target
intercept~\cite{Beard02}, terrain mapping~\cite{Marco03},
reconnaissance~\cite{Ous04a}, and surveillance~\cite{Stoeter02}. To
achieve effective solutions, in general, a vehicle team needs to
follow a cooperative policy. The generation of such a policy has been
the subject of a rich literature in cooperative control. A sample of
the noteworthy work in this field includes a language for modeling and
programming cooperative control systems~\cite{klavins-icra04},
receding horizon control for multi-vehicle systems~\cite{dunbar04},
non-communicative multi-robot coordination~\cite{Kok05}, hierarchical
methods for target assignment and intercept~\cite{Beard02},
cooperative estimation for reconnaissance problems~\cite{Ous04a},
mixed integer linear programming methods for cooperative
control~\cite{Schouwenaars01,Earl02}, the compilation on  multi-robots
in dynamics environments~\cite{Lima05}, and the compilation on
cooperative control and optimization~\cite{Murphey02}.  

When multi-vehicle teams operate in dynamically changing and uncertain
environments, which is often the case, a model predictive
approach~\cite{Mayne00} can be used to provide feedback. This approach
involves frequently recomputing the team control policy in real-time.
However, because these systems are often hybrid dynamical systems,
computing a cooperative policy is often computationally hard. The
challenge addressed in this paper is to (1) develop a method to
generate near-optimal cooperative policies quickly and (2) to
effectively implement the method.

In our previous work on cooperative control~\cite{Earl02,Earl04a} we
developed mixed integer linear programming methods because of their
expressiveness and ease of modeling many types of problems. The
drawback is that real-time planning is infeasible because of the
computational complexity of the approach. This motivated us to develop
a trajectory primitive decomposition approach to the problem. This
approach finds near-optimal solutions quickly, allowing real-time
implementation, and can be tuned to balance the tradeoff between
optimality and computational effort for the particular problem at
hand. The drawback, compared to our previous work, is that it is
limited to cooperative control problems in which vehicle tasks can be
clearly defined and efficient primitives exist.

In this paper, we present our trajectory primitive decomposition
approach. We analyze the average case behavior of the approach by
solving instances of a cooperative control problem derived from
Cornell's RoboFlag environment. And finally, we implement the approach
in a hierarchical architecture with variable replanning rates at each
level and test the implementation in a dynamically changing and
uncertain RoboFlag environment.

The trajectory primitive decomposition involves the introduction of a
set of tasks to be executed by the vehicles, allowing the problem to
be separated into a low-level component, called task execution, and a
high-level component, called task assignment.  The task execution
component is formulated as an optimal control problem, which
explicitly involves the vehicle dynamics. Given a vehicle and a task,
the goal is to find the control inputs necessary to execute the given
task in an optimal way. The task assignment component is an
NP-hard~\cite{Garey} combinatorial optimization problem. The goal
is to assign a sequence of tasks to each vehicle so that the team
objective is optimized. Task assignment does not explicitly involve
the vehicle dynamics because the task execution component is utilized
as a trajectory primitive.

We have developed a branch and bound algorithm to solve the task assignment
problem. One of the benefits of this algorithm is that it can be
stopped at any time in the solution process and the output is the
best feasible assignment found in that time. This is advantageous  for
real-time applications where control strategies must be generated
within a time window. In this case, the best solution found in the
time window is used. Another advantage is that the algorithm is
complete; given enough time, it will find the optimal solution.

To analyze the average case performance of the branch and bound
solver, we generate and solve many instances of the problem. We look
at computational complexity, convergence to the optimal assignment,
and performance variations with parameter changes.  We found that the
solver converges to the optimal assignment quickly.  However, the
solver takes much more time to prove the assignment is optimal.
Therefore, if the solver is terminated early, the best feasible
assignment found in that time is likely to be a good one.  We also
found several phase transitions in the task assignment problem,
similar to those found in the pioneering work~\cite{kirk94,selman,Monasson}.
At the phase transition point, the task assignment problem is much
harder to solve.  For cooperative control problems involving
adversaries, the transition point occurs when the capabilities of the
two teams are comparable.  This behavior is similar to the complexity
of balanced games like chess~\cite{Herik}.

Finally, we implement the methods in a multi-level architecture with
replanning occurring at each level, at different rates (multi-level
model predictive control). The motivation is to provide feedback to
help handle dynamically changing environments.

The paper is  organized as follows: In Section~\ref{sec:mvta}, we
state the multi-vehicle cooperative control problem and introduce the
decomposition. In Section~\ref{sec:roboflag}, we introduce the example
problem used to motivate our approach. In Section~\ref{sec:bb}, we
describe our solver for the task assignment problem, and in
Section~\ref{sec:analysis}, we analyze its average case behavior.
Finally, in Section~\ref{sec:mpc}, we apply our solver in a
dynamically changing and uncertain environment using a multi-level
model predictive control architecture for feedback.  A web page that
accompanies this paper can be found at~\cite{EarlWebPage}.

\section{Multi-vehicle task assignment}
\label{sec:mvta}
The general multi-vehicle cooperative control problem consists of a
heterogeneous set of vehicles (the team), an operating environment,
operating constraints, and an objective function. The goal is to
generate a team strategy that minimizes the objective function.  The
strategy in its lowest level form is the control inputs to each
vehicle of the team.

In~\cite{Earl04a,Earl02}, we show how to solve this problem using
hybrid systems tools. This approach is successful in determining
optimal strategies for complex multi-vehicle problems, but becomes
computationally intensive for large problems. Motivated to find faster
techniques, we have developed a decomposition approach described in
this paper. 

The key to the decomposition is to introduce a relevant set of tasks
for the problem being considered.  Using these tasks, the problem can
be decomposed into a task completion component and a task assignment
component.  The task completion component is a low level problem,
which involves a vehicle and a task to be completed.  The task
assignment component is a high level problem, which involves the
assignment of a sequence of tasks to be completed by each vehicle in
the team.  

\emph{Task Completion}:
Given a vehicle, an operating environment with constraints, a
task to be completed, and an objective function, find the control
inputs to the vehicle such that the constraints are satisfied, the
task is completed, and the objective is minimized. 

\emph{Task Assignment}: Given a set of vehicles, a task completion
algorithm for each vehicle, a set
of tasks to be completed, and an objective function, assign a sequence
of tasks to each vehicle such that the objective function is
minimized.

\begin{figure}
\centering
\includegraphics[width=200pt]{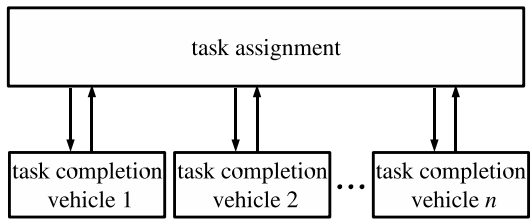}
\caption{The framework for the task assignment problem using task 
completion primitives. 
}
\label{fig:taframe}
\end{figure}

In the task assignment problem, instead of  varying the control inputs
to the vehicles to find an optimal strategy, we vary the sequence of
tasks assigned to each vehicle.  This problem is a combinatorial
optimization problem and does not explicitly involve the dynamics of
the vehicles.  However, in order to calculate the objective function
for any particular assignment, we must use the task completion
algorithm. Task completion acts as a primitive in solving the task
assignment problem, as shown by the framework in
Figure~\ref{fig:taframe}.
Using the low level component (task completion),
the high level component (task assignment) need not
explicitly consider the detailed dynamics of the vehicles required to
perform a task.

\section{RoboFlag Drill}
\label{sec:roboflag}
\begin{figure}
\centering
\includegraphics[width=3.0in]{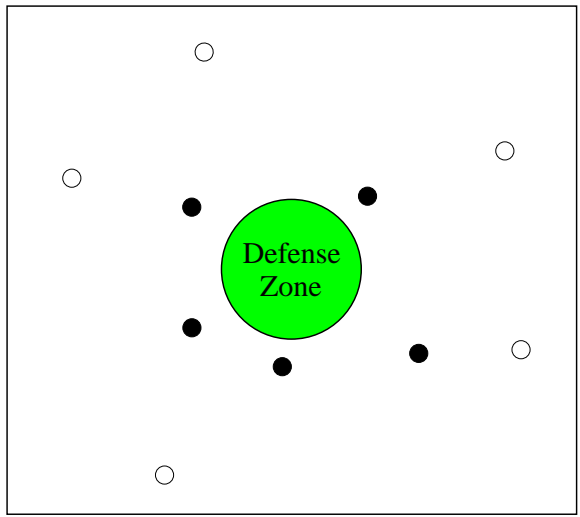}
\caption{The RoboFlag Drill used to 
motivate the methods presented is this paper.
The drill takes place on a playing field with a Defense Zone at its
center. The objective is to design a cooperative control strategy for
the team of defending vehicles (black) that minimizes the number of
attacking vehicles (white) that enter the Defense Zone.}
\label{fig:drill}
\end{figure}
To motivate and make concrete our decomposition approach, we
illustrate the approach on an example problem derived from Cornell's
multi-vehicle system called RoboFlag. For an introduction to RoboFlag,
see the papers from the invited session on RoboFlag in the Proceedings
of the 2003 American Control
Conference~\cite{campbell03,d'andrea03a,d'andrea03b}.
In~\cite{klavins-cdc03-ccl}, protocols for the RoboFlag Drill are
analyzed using a computation and control language.

The RoboFlag Drill involves two teams of vehicles, the defenders and
the attackers, on a playing field with a circular region of radius
$R_{dz}$ at its center called the Defense Zone
(Figure~\ref{fig:drill}).   The attackers' objective is to fill the
Defense Zone with as many attackers as possible.  They have a fixed
strategy in which each moves toward the Defense Zone at constant
velocity.  An attacker stops if it is intercepted by a defender or if
it enters the Defense Zone.  The defenders' objective is to deny as
many attackers as possible from entering the Defense Zone without
entering the zone themselves. A defender denies an attacker from the
Defense Zone by intercepting the attacker before it reaches the
Defense Zone. 

The wheeled vehicles of Cornell's RoboCup
Team~\cite{Stone01,d'andrea01} are the defenders in the RoboFlag Drill
problem we consider in this paper.  Each vehicle is equipped with a
three-motor omni-directional drive that allows it to  move along any
direction irrespective of its orientation.  This allows for superior
maneuverability compared to traditional nonholonomic (car-like)
vehicles.  A local control system on the vehicle, presented
in~\cite{Nagy04} and Appendix~\ref{sec:vehicleDynamics}, alters the
dynamics so that at a higher level of the hierarchy, the vehicle
dynamics are governed by
\begin{eqnarray}
&&\ddot{x}(t) + \dot{x}(t) = u_x(t)\nonumber\\%
&&\ddot{y}(t) + \dot{y}(t) = u_y(t)\nonumber\\%
&&u_x(t)^2 + u_y(t)^2 \leq 1.
\label{eom1}
\end{eqnarray}
The state vector is  $\mathbf{x} = (x,y,\dot{x},\dot{y})$, and the
control input vector is $\mathbf{u} = (u_x,u_y)$.  These equations are
less complex than the nonlinear governing equations of the vehicles.  They
allow for the generation of feasible near-optimal trajectories with
little computational effort and have been used successfully in the
RoboCup competition.

\begin{figure}
\centering
\includegraphics[width=2.5in]{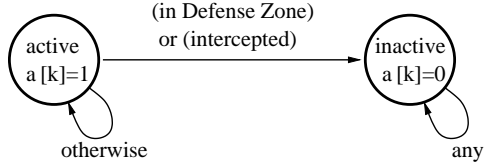}
\caption{The two state (active and inactive) attacker state machine.
The attacker starts in the active state. It transitions to the
inactive state, and remains in this state, if it enters the Defense
Zone or if it is intercepted by a defender.}
\label{sm1}
\end{figure}

Each attacker has two discrete modes: active and inactive.  When
active, the attacker moves toward the Defense Zone at constant
velocity along a straight line path.  The attacker, which is initially 
active, transitions to inactive mode if
the defender intercepts it or if it enters the Defense Zone.   Once
inactive, the attacker does not move and remains inactive for the
remainder of play.  These dynamics are captured by the 
discrete time equations 
\begin{eqnarray}
  &&p[k+1] = p[k] + v_p T a[k]\nonumber\\
  &&q[k+1] = q[k] + v_q T a[k]
  \label{attdyn1}
\end{eqnarray}
and the state machine (see Figure~\ref{sm1})
\begin{eqnarray}
  \label{attdyn2}
  &&a[k+1] = \left\{
  \begin{array}{ll}
    1 & \mbox{if ($a[k]=1$)} \\
      & \mbox{and (not in Defense Zone)} \\
      & \mbox{and (not intercepted) } \\
    0 & \mbox{if ($a[k]=0$)}\\
      & \mbox{or (in Defense Zone)} \\
      & \mbox{or (intercepted)}
  \end{array} \right. 
\end{eqnarray}
for all $k$ in the set $\{ 1,\ldots,N_a \}$.  
The initial conditions are
\begin{eqnarray}
  \mbox{$p[0] = p_s$, $q[0] = q_s$, and $a[0] = 1$}.
  \label{attdyn3}
\end{eqnarray}
In these equations, $N_a$ is the number of samples, $T$ is the sample
time, $(p[k],q[k])$ is the attacker's position at time $t_a[k] = k T$,
$(v_p,v_q)$ is its constant velocity vector, and $a[k] \in \{0,1 \}$
is a discrete state indicating the attacker's mode.  The attacker is
active when $a[k] = 1$ and inactive when $a[k] = 0$.  Given
$(p[k],q[k])$ and $a[k]$, we can calculate the attacker's position at
any time $t$, denoted $\mathbf{p}(t)=(p(t),q(t))$, using the
equations
\setlength{\arraycolsep}{0.0em}
\begin{eqnarray}
p(t) &{}={}& p[k] + v_p a[k] (t-t_a[k])\nonumber\\
q(t) &{}={}& q[k] + v_q a[k] (t-t_a[k]),
\label{eqn:inbetween}
\end{eqnarray}
\setlength{\arraycolsep}{5pt}%
where $k = \lfloor t/T \rfloor$.

Because the goal of the RoboFlag Drill is to keep attackers out of the
Defense Zone, attacker intercept is an obvious task for this problem.
Therefore, the task completion problem for the RoboFlag Drill is an
intercept problem. 

\emph{RoboFlag Drill Attacker Intercept (RDAI)}:
Given a defender with state $\mathbf{x}(t)$ governed by
equation~(\ref{eom1}) with initial condition
$\mathbf{x}(0)=\mathbf{x}_s$, an attacker governed by
equations~(\ref{attdyn1}) and~(\ref{attdyn2}) with initial conditions
given by~(\ref{attdyn3}) and coordinates $\mathbf{p}(t)$ given by
equation~(\ref{eqn:inbetween}), obstacles and restricted regions to be
avoided, time dependent final condition
$\mathbf{x}(t_f)=(p(t_f),q(t_f),0,0)$, and objective function
$J_{TC}=t_f$, find the control inputs to the defender that minimize
the objective such that the constraints are satisfied.

The operating environment includes the playing field and the group of
attacking vehicles.  The operating constraints include collision
avoidance between vehicles and avoidance of the Defense Zone (for the
defending robots).  

Next, we define notation for a primitive that generates a trajectory
solving the RDAI problem.  The inputs to the primitive are the current
state of defender $d$, denoted $\mathbf{x}_d(t)$, and the current
position of attacker $a$, denoted $\mathbf{p}_a(t)$.  The output is
the amount of time it takes defender $d$ to intercept attacker $a$,
denoted $\Delta t_{int}(d,a,t)$, given by
\begin{eqnarray} 
\Delta t_{int}(d,a,t) := {\tt intTime}[\mathbf{x}_d(t),\mathbf{p}_a(t)  ].  
\label{eqn:primitive}
\end{eqnarray} 
If defender $d$ can not intercept attacker $a$ before the attacker
enters the Defense Zone, we set $\Delta t_{int}(d,a,t) := \infty$.

Near-optimal solutions to the RDAI problem can be generated using the
technique presented in~\cite{Nagy04} with straightforward
modification. The advantage of this technique is that it finds very
good solutions quickly, which allows for the exploration of many
trajectories in the planning process. Another way to generate
near-optimal solutions for RDAI is to use the iterative mixed integer
linear programming techniques presented in~\cite{Earl04b}.  The
advantage of this approach is that it can handle complex hybrid
dynamical systems. Either of these approaches could be used as a
primitive for the RDAI problem.  Using the primitive, the RoboFlag
Drill problem can be expressed as the following task assignment
problem:

\emph{RoboFlag Drill Task Assignment (RDTA)}: Given a team of
defending vehicles $\mathcal{D}=\{ d_1,\ldots,d_n \}$, a set of
attackers to intercept $\mathcal{A}=\{ a_1,\ldots,a_m \}$, initial
conditions for each defender $d$ and for each attacker $a$, an RDAI
primitive, and an objective function $J$, assign each defender $d$ in
$\mathcal{D}$ a sequence of attackers to intercept, denoted
$\alpha_d$, such that the objective function is minimized.

\begin{table}
\begin{center}
\caption{Variables for RoboFlag Drill problems}
\begin{tabular}{|l|l|}
\hline
$n$  & number of defending vehicles  \\ \hline 
$m$ & number of attacking vehicles   \\ \hline
$\mathcal{D}$ & the set of defending vehicles   \\ \hline
$\mathcal{A}$ & the set of attacking vehicles   \\ \hline
$\mathcal{A}_u$ & the set of unassigned attacking vehicles   \\ \hline
$\mathcal{A}-\mathcal{A}_u$ & the set of assigned attacking vehicles   \\ \hline
$\mathbf{x}_d(t)$ & the state of defender $d$ at time $t$  \\ \hline
$\mathbf{p}_a(t)$ & the position of attacker $a$ at time $t$  \\ \hline
$\alpha_d$ & the sequence of attackers assigned to defender $d$  \\ \hline
$m_d$ & the length of defender $d$'s intercept sequence $\alpha_d$  \\ \hline
$\Delta t_{int}(d,a,t)$ & time needed for $d$ to intercept $a$
starting at time $t$.  \\ \hline
$t_d(i)$ & 
the time that $d$ completes $i$th task in task sequence
$\alpha_d$  \\ \hline
$\gamma_a$ & binary variable indicating if $a$ enters the Defense Zone  \\ \hline
$J$ & the cost function for the RDTA problem  \\ \hline
$\epsilon$ & weight in the cost function $J$  \\ \hline
\end{tabular}
\label{tbl:variables}
\end{center}
\end{table}

We introduce notation 
(listed in Table~\ref{tbl:variables}) to 
describe the cost function $J$ and the
algorithm that solves the RDTA problem.
Let $m_d$ be the number of
attackers assigned to defender $d$, and let $\alpha_d=\langle
\alpha_d(1),\ldots,\alpha_d(m_d)\rangle$ be the sequence of attackers
defender $d$ is assigned to intercept.  Let $t_d(i)$ be the time at
which defender $d$ completes the $i$th task in its task sequence
$\alpha_d$.
Let $\mathcal{A}_u$ be the set of unassigned attackers, then
$\mathcal{A} - \mathcal{A}_u$ is the set of assigned attackers.

An assignment for the RDTA problem is an intercept sequence $\alpha_d$
for each defender $d$ in $\mathcal{D}$. A partial assignment is an
assignment such that $\mathcal{A}_u$ is not empty, and a complete
assignment is an assignment such that $\mathcal{A}_u$ is empty.

The set of times $\{ t_d(i) : i=1,\ldots,m_d\}$, for each defender
$d$, are computed using the primitive in
equation~(\ref{eqn:primitive}).  The time at which defender $d$
intercepts the $i$th attacker in its intercept sequence, if not empty,
is given by
\begin{eqnarray}
t_d(i) = 
\left\{ 
\begin{array}{l}
t_d(i-1), \mbox{ if } \Delta t_{int}(d,\alpha_d(i),t_d(i-1)) = \infty \\
t_d(i-1) + \Delta t_{int}(d,\alpha_d(i),t_d(i-1)), \mbox{
otherwise}\nonumber
\end{array},
\right.
\end{eqnarray}
where we take $t_d(0) = 0$.  If defender $d$ can not intercept
attacker $\alpha_d(i)$ before the attacker enters the Defense Zone,
the time $t_d(i)$ is not incremented because, in this case, the
defender does not attempt to intercept the attacker.  The time at
which defender $d$ completes its intercept sequence $\alpha_d$ is
given by $t_d(m_d)$.

To indicate if attacker $a$ enters the Defense Zone during the drill,
we introduce binary variable $\gamma_a$ given by
\begin{eqnarray}
\gamma_a = \left\{ 
\begin{array}{ll} 
1 & \mbox{if attacker $a$ enters Defense Zone}\\  
0 & \mbox{otherwise}.   
\end{array}
\right.
\end{eqnarray}
If $\gamma_a=1$, attacker $a$ enters the Defense Zone at some time
during play, otherwise, $\gamma_a=0$ and attacker $a$ is intercepted.
We compute $\gamma_a$ for each attacker $a$ in the set of assigned
attackers ($\mathcal{A} - \mathcal{A}_u)$ as follows:  For each $d$ in
$\mathcal{D}$ and for each $i$ in $\{1,\ldots,m_d\}$, if $\Delta
t_{int}(d,\alpha_d(i),t_d(i-1)) = \infty$ then set
$\gamma_{\alpha_d(i)} = 1$, otherwise set $\gamma_{\alpha_d(i)} = 0$.

For the RDTA problem, the cost function has two components.  The
primary component is the number of assigned attackers that enter the
Defense Zone during the drill,
\begin{eqnarray}
J_1 = \sum_{a \in (\mathcal{A}-\mathcal{A}_u)} \gamma_a.
\end{eqnarray}
The secondary component is the time at which all assigned attackers
that do not enter the Defense Zone (all $a$ such that $\gamma_a = 0$)
are intercepted,  
\begin{eqnarray}
J_2 = \max_{d \in \mathcal{D}} t_d(m_d).
\end{eqnarray}
The weighted combination is
\begin{eqnarray}
\label{eqn:objective}
J = \sum_{a \in (\mathcal{A}-\mathcal{A}_u)} \gamma_a + 
\epsilon\max_{d \in \mathcal{D}} t_d(m_d),
\end{eqnarray}
where we take $0< \epsilon \ll 1$ because we want the primary
component to dominate.  In particular, keeping attackers out of the
Defense Zone is most important. Therefore, our goal in the RDTA
problem is to generate a complete assignment ($\mathcal{A}_u$ empty)
that minimizes equation~(\ref{eqn:objective}).

\section{Branch and bound solver}
\label{sec:bb}
One way to find the optimal assignment for RDTA is by exhaustive
search; try every possible assignment of tasks to vehicles and pick
the one that minimizes $J$.  This approach quickly becomes
computationally infeasible for large problems.  As the number of tasks
or the number of vehicles increase, the total number of possible
assignments grows significantly.  A more efficient solution method is
needed for real-time planning.  With this motivation, we developed a
branch and bound solver for the problem.  In this section, we describe
the solver and its four major components: node expansion, branching,
upper bound, and lower bound.

We use a search tree to enumerate all possible assignments for the
problem. The root node represents the empty assignment, all interior
nodes represent partial assignments, and the leaves represent the set
of all possible complete assignments.  Given a node representing a
partial assignment, the node expansion algorithm
(Section~\ref{sec:nodeExpand}) generates the node's children.  Using
the node expansion algorithm, we grow the search tree starting from
the root node. The branching algorithm (Section~\ref{sec:Branch}) is
used to determine the order in which nodes are expanded. In this
algorithm, we use A* search~\cite{stefik} to guide the growth of the
tree toward good solutions.

Given a node in the tree representing a partial assignment, the
upper bound algorithm (Section~\ref{sec:ubAlgo}) assigns the
unassigned attackers in a greedy way.  The result is a feasible
assignment. The cost of this assignment is an upper bound on the
optimal cost that can be achieved from the given node's partial
assignment. The upper bound is computed at each node explored in the
tree (not all nodes are explored, many are pruned).  As
the tree is explored, the best upper bound found to date is stored in
memory.

Given a node in the search tree representing a partial assignment, the
lower bound algorithm (Section~\ref{sec:lbAlgo}) assigns the
unassigned attackers in $\mathcal{A}$ using the principle of
simultaneity. Each defender is allowed to pursue multiple attackers
simultaneously. Because this is physically impossible, the resulting
assignment is potentially infeasible.  Because no feasible assignment
can do better, the cost of this assignment is a lower bound on the
cost that can be achieved from the given node's partial assignment.
Similar to the upper bound, the lower bound is computed at each node
explored in the tree. 

If the lower bound  for the current node being
explored is greater or equal to the best upper bound found, we prune
the node from the tree, eliminating all nodes that emanate from the
current node.  This can be done because of the way we have constructed
the tree. The task sequences that make up a parent's assignment are
subsequences of the sequences that make up each child's assignment.
Therefore, exploring the descendants will not result in a better
assignment than that already obtained.

\begin{table}
\caption{Branch and bound algorithm}
\label{tbl:BandB}
\begin{center}
\framebox{\parbox{3.4in}{
\begin{algorithmic}[1]
\STATE  Start with a tree containing only the root node. 
\STATE  Run \emph{upper bound} algorithm with root node's partial
assignment (the empty assignment) as input, generating a feasible
complete assignment.
\STATE  Set $J_{ub}^{best}$ to the cost the complete assignment. 
\STATE  Expand the root node using \emph{expand node} routine. 
\WHILE {growing the tree}
\STATE Use \emph{branching} routine to pick next branch to explore.
\STATE Use \emph{upper bound} algorithm to compute feasible complete
assignment from current node's partial assignment, and set the cost of
this assignment to $J_{ub}$.
\STATE if $J_{ub} < J_{ub}^{best}$, set $J_{ub}^{best} := J_{ub}$.
\STATE Use \emph{lower bound} algorithm to calculate the lower bound 
cost from the
current node's partial assignment, and set this cost to  $J_{lb}$.  
\STATE if $J_{lb} \geq J_{ub}^{best}$, prune current node from the tree.
\ENDWHILE
\end{algorithmic}
}}
\end{center}
\end{table}

Before we describe the details of the components, we describe
the branch and bound algorithm listed in Table~\ref{tbl:BandB}.  Start
with the root node, which represents the empty assignment, and apply
the upper bound algorithm.  This generates a feasible assignment
with cost denoted $J_{ub}^{best}$ because it is the best, and only,
feasible solution generated so far. Next, apply the node expansion
algorithm to root, generating its children. 

At this point, enter a loop. For each iteration of the loop, apply the
branching algorithm to select the node to explore next.  The node
selected by the branching algorithm, which we call the current node,
contains a partial assignment.  Apply the upper bound algorithm to the
current node, generating a feasible complete assignment with cost
denoted $J_{ub}$.  If $J_{ub}$ is less than $J_{ub}^{best}$, we have
found a better feasible assignment so we set $J_{ub}^{best} :=
J_{ub}$.  Next, apply the lower bound algorithm to generate an
optimistic cost, denoted $J_{lb}$, from the current node's partial
assignment.  If $J_{lb}$ is greater than or equal to the best feasible
cost found so far $J_{ub}^{best}$, prune the node from the tree,
removing all of its descendants from the search.  We do not need
to consider the descendants of this node because doing so will not
result in a better feasible assignment than the one found already,
with cost $J_{ub}^{best}$.  The loop continues until all nodes have
been explored or pruned away.  The result is the optimal assignment
for the RDTA problem.

\begin{figure}
\centering
\includegraphics[width=250pt]{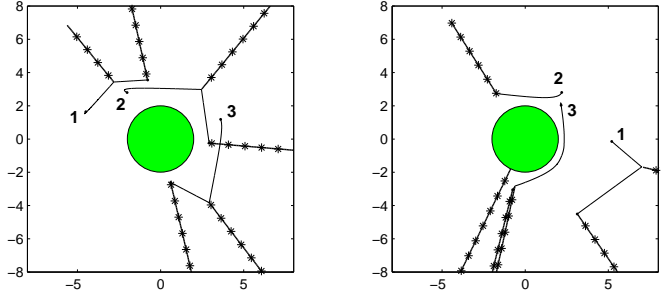}
\caption{The solution to two instances of the RDTA problem using the
branch and bound solver.  The circle at the center of the field is the
Defense Zone.  The lines with asterisks denote the attacker
trajectories, and the lines without denote defender trajectories.  The
parameters for these instances are $\epsilon=0.01$, $n=3$, and $m=6$. 
}
\label{fig:4now}
\end{figure}

In Figure~\ref{fig:4now}, we plot the solution to two instances of the RDTA
problem solved using the branch and bound solver. Notice that the
defenders work together and do not greedily pursue the attackers that
are closest. For example, in the figure on the left, defenders 2 and 3 
ignore the closest attackers and pursue attackers further away
for the benefit of the team.

In the remainder of this section we describe the components of the
branch and bound solver in detail.

\subsection{Node expansion} 
\label{sec:nodeExpand}

Here we describe the node expansion algorithm  used to grow a search
tree that enumerates all possible assignments for the RDTA problem.
Each node of the tree represents an assignment. Starting from the root
node, attackers are assigned, forming new nodes, until all complete
assignments are generated. Each node represents a partial
assignment except for the leaves, which represent the set of complete
assignments.

\begin{figure}
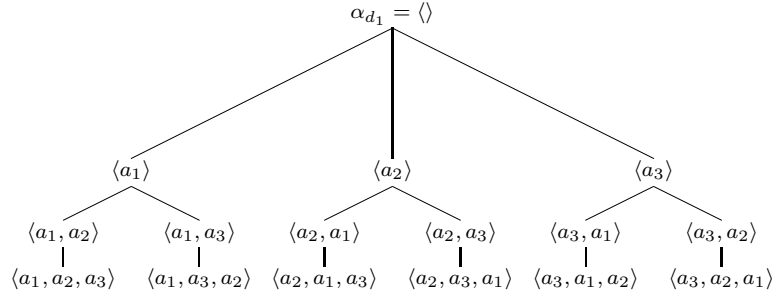

\begin{center}
{\footnotesize
\qtreecenterfalse
\Tree 
[.{$\alpha_{d_1}=\langle\rangle$} 
  [.{$\langle a_1 \rangle$} 
    [.{$\langle a_1,a_2 \rangle$} 
      {$\langle a_1,a_2,a_3 \rangle$} 
    ]
    !\qsetw{.7in} 
    [.{$\langle a_1,a_3 \rangle$} 
      {$\langle a_1,a_3,a_2 \rangle$} 
    ]
  ]
  !\qsetw{1.30in} 
  [.{$\langle a_2 \rangle$} 
    [.{$\langle a_2,a_1 \rangle$} 
      {$\langle a_2,a_1,a_3 \rangle$} 
    ]
    !\qsetw{.7in} 
    [.{$\langle a_2,a_3 \rangle$} 
      {$\langle a_2,a_3,a_1 \rangle$} 
    ]
  ] 
  !\qsetw{1.3in} 
  [.{$\langle a_3 \rangle$}
    [.{$\langle a_3,a_1 \rangle$} 
      {$\langle a_3,a_1,a_2 \rangle$} 
    ]
    !\qsetw{.7in} 
    [.{$\langle a_3,a_2 \rangle$} 
      {$\langle a_3,a_2,a_1 \rangle$} 
    ]
  ] 
]
}
\end{center}
\caption{Search tree for the RDTA problem with the
defender set $\mathcal{D} = \{ d_1 \}$ and the attacker set
$\mathcal{A}=\{a_1,a_2,a_3 \}$. Each node of the tree denotes a
sequence of attackers to be intercepted by defender $d_1$. The root
node is the empty assignment. The leaves of the tree give all possible
complete assignments of attackers in $\mathcal{A}$.}
\label{simpleTree}
\end{figure}

Consider the case with one defender $\mathcal{D}=\{ d_1 \}$ and three
attackers $\mathcal{A}=\{ a_1, a_2,  a_3 \}$. The tree for this case
is shown in Figure~\ref{simpleTree}. To generate this tree, we start
from the root node representing the empty assignment, denoted $\langle
\rangle$.  We expand the root node generating three children, each
representing an assignment containing a single attacker to intercept.
The children are then expanded, and so on, until all possible
assignments are generated.

\begin{figure}
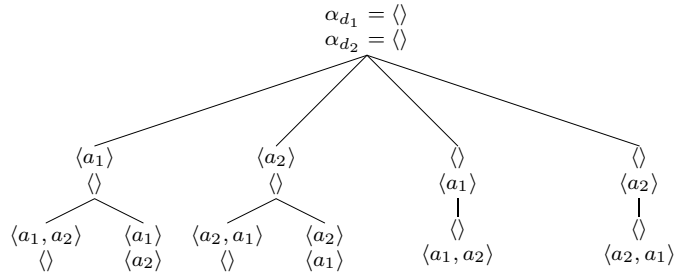

\begin{center}
{\footnotesize
\qtreecenterfalse
\Tree
[.{$\begin{array}{c}
\alpha_{d_1}=\langle\rangle \\ 
\alpha_{d_2}=\langle\rangle \end{array}$}
  [.{$\begin{array}{c} \langle a_1 \rangle \\ \langle \rangle \end{array}$}
    {$\begin{array}{c} \langle a_1,a_2 \rangle \\ \langle \rangle \end{array}$}
    !\qsetw{0.5in}
    {$\begin{array}{c} \langle a_1 \rangle \\ \langle a_2 \rangle \end{array}$}
  ]
  !\qsetw{1.0in}
  [.{$\begin{array}{c} \langle a_2 \rangle \\ \langle \rangle \end{array}$}
    {$\begin{array}{c} \langle a_2,a_1 \rangle \\ \langle \rangle \end{array}$}
    !\qsetw{0.5in}
    {$\begin{array}{c} \langle a_2 \rangle \\ \langle a_1 \rangle \end{array}$}
  ]
  !\qsetw{0.9in}
  [.{$\begin{array}{c} \langle \rangle \\ \langle a_1 \rangle \end{array}$}
    {$\begin{array}{c} \langle \rangle \\ \langle a_1,a_2 \rangle \end{array}$}
  ]
  !\qsetw{0.9in}
  [.{$\begin{array}{c} \langle \rangle \\ \langle a_2 \rangle \end{array}$}
    {$\begin{array}{c} \langle \rangle \\ \langle a_2,a_1 \rangle \end{array}$}
  ]
]
}
\end{center}
\caption{Search tree for the RDTA problem with defender 
set $\mathcal{D} = \{ d_1, d_2 \}$ and the attacker set
$\mathcal{A}=\{a_1,a_2 \}$. Each node of the tree denotes a sequence
of attackers to intercept for defender $d_1$ and defender $d_2$. The
root node is the empty assignment. The leaves of the tree give all
possible complete assignments of attackers in $\mathcal{A}$.}
\label{simpleTree2}
\end{figure}

For multiple defenders, unlike the single defender case, the tree is
unbalanced to avoid redundancies. For example, consider the case with
two defenders $\mathcal{D} = \{ d_1, d_2\}$ and two attackers
$\mathcal{A}=\{ a_1, a_2 \}$. The tree for this case is shown in
Figure~\ref{simpleTree2}.  Again, each node represents an assignment,
but now the assignment is a sequence of attackers to intercept for
each defender in $\mathcal{D}$.
In general, for defender set $\mathcal{D}$ with $n$ defenders and
attacker set $\mathcal{A}$ with $m$ attackers
there are $(n+m-1)!/(n-1)!$
complete assignments (or leaves in the search tree).

To generate a search tree for the general case, we use a node
expansion algorithm.  This algorithm takes any node and generates the
node's children.  The assignment for each child is constructed by
appending an unassigned attacker to one of the sequences in the parent
node's assignment.  The task sequences in the parent's assignment
are always subsequences of the sequences in its child's
assignment. Therefore, when we prune a node from the search tree, we
can also prune all of its descendants. 

The node expansion algorithm uses a different representation for an
assignment than we have used thus far. We introduce this new
representation with an example involving the defender set
$\mathcal{D}=\{ d_1,d_2\}$ and the attacker set
$\mathcal{A}=\{a_1,a_2,\ldots,a_7 \}$.  Consider a partial assignment
given by 
\begin{eqnarray}
  &&\alpha_{d_1} = \langle a_4,a_1 \rangle\nonumber\\
  &&\alpha_{d_2} = \langle a_2,a_5,a_7 \rangle\nonumber.
\end{eqnarray}
In this case, attackers $a_3$ and $a_6$ have yet to be assigned.
Our node expansion algorithm represents this partial assignment with 
the vectors 
\begin{eqnarray}
&&\mathbf{\delta} = (1,1,2,2,2,0,0)\nonumber\\%
&&\mathbf{\beta} = (4,1,2,5,7,0,0),%
\label{expandExample}
\end{eqnarray}
both of length $m = 7$.  Vector $\mathbf{\delta}$ holds defender
indices and vector $\mathbf{\beta}$ holds attacker indices.  For a
unique representation, the elements in $\mathbf{\delta}$ are ordered
so that $\delta(i) \leq \delta(i+1)$. For the example case, attackers
$a_{\beta(1)}$ and $a_{\beta(2)}$ (i.e.,~$a_4$ and $a_1$) are assigned to
defender $d_1$ in sequence, and attackers $a_{\beta(3)}$,
$a_{\beta(4)}$, $a_{\beta(5)}$ (i.e.,~$a_2$, $a_5$, $a_7$) are assigned to
defender $d_{2}$ in sequence. 

In general, the input to the node expansion algorithm is a parent node with
assignment give by
\begin{eqnarray}
&& \mbox{parent}.\mathbf{\delta} = 
(\delta(1),\delta(2),\ldots,\delta(p),0,\ldots,0)\nonumber\\
&& \mbox{parent}.\mathbf{\beta} = 
(\beta(1),\beta(2),\ldots,\beta(p),0,\ldots,0),
\end{eqnarray}
where both vectors are of size $m$, and $p$ is the number of
tasks already assigned (or the number of nonzero entries in each
vector). The output is a set of $N_{child}$ children, where  
\begin{eqnarray}
N_{child} = (n - \delta(p) + 1) 
(m-p).
\end{eqnarray}
Each child  has assignment vectors $\mathbf{\delta}$ and
$\mathbf{\beta}$ identical to its parent except for entries
$\delta(p+1)$ and $\beta(p+1)$. In the child's assignment, attacker
$a_{\beta(p+1)}$ is appended to defender $d_{\delta(p+1)}$'s sequence of
attackers to intercept $\alpha_{\delta(p+1)}$. The details of the node
expansion algorithm are given in Table~\ref{tbl:nodeExpand}. 
\begin{table}
\caption{Node expansion algorithm}
\label{tbl:nodeExpand}
\begin{center}
\framebox{ \parbox{3.4in}{
\begin{algorithmic}[1]
\STATE  { $k := 1$ } 
\FOR {$i=\delta(p),\delta(p+1),\ldots,n$}
  \FOR {each $j$ in the set 
        $\{\{1,\ldots,m \} - \{\beta(1),\beta(2),\ldots,\beta(p)\}\}$ }
    \STATE { $\mbox{child}(k).\mathbf{\delta} =
             \mbox{parent}.\mathbf{\delta}$ }
    \STATE { $\mbox{child}(k).\mathbf{\beta} =
             \mbox{parent}.\mathbf{\beta}$ }  
    \STATE $\mbox{child}(k).\delta(p+1) = i$
    \STATE $\mbox{child}(k).\beta(p+1) = j$
    \STATE $k := k + 1$
  \ENDFOR
\ENDFOR
\end{algorithmic}
}}
\end{center}
\end{table}

\begin{figure}
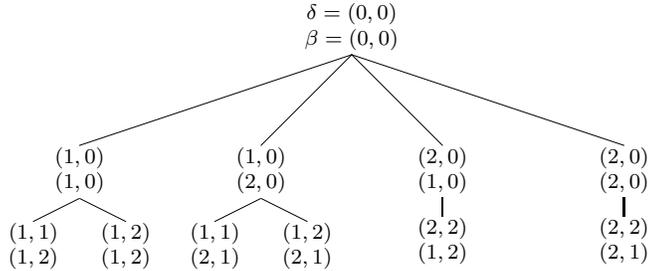

\begin{center}
{\footnotesize
\qtreecenterfalse
\Tree
[.{$\begin{array}{c} {\bf \delta} = (0,0) \\ {\bf \beta} = (0,0) \end{array}$}
  [.{$\begin{array}{c} (1, 0) \\ (1, 0) \end{array}$}
    {$\begin{array}{c} (1, 1) \\ (1, 2) \end{array}$}
    !\qsetw{0.4in}
    {$\begin{array}{c} (1, 2) \\ (1, 2) \end{array}$}
  ]
  !\qsetw{1.0in}
  [.{$\begin{array}{c} (1, 0) \\ (2, 0) \end{array}$}
    {$\begin{array}{c} (1, 1) \\ (2, 1) \end{array}$}
    !\qsetw{0.4in}
    {$\begin{array}{c} (1, 2) \\ (2, 1) \end{array}$}
  ]
  !\qsetw{0.9in}
  [.{$\begin{array}{c} (2, 0) \\ (1, 0) \end{array}$}
    {$\begin{array}{c} (2, 2) \\ (1, 2) \end{array}$}
  ]
  !\qsetw{0.9in}
  [.{$\begin{array}{c} (2, 0) \\ (2, 0) \end{array}$}
    {$\begin{array}{c} (2, 2) \\ (2, 1) \end{array}$}
  ]
]
}
\end{center}
\caption{The tree from Figure~\ref{simpleTree2} written using our
node expansion algorithm's notation.}
\label{fig:nodeExpand3}
\end{figure}

To demonstrate the node expansion algorithm, we expand the node given by
equation~(\ref{expandExample}) as shown in Figure~\ref{fig:nodeExpand}.  
Figure~\ref{fig:nodeExpand2} shows the normal notation for this expansion.
\begin{figure}
\begin{center}
{\footnotesize
\qtreecenterfalse
\Tree
[.{$\begin{array}{c}
   \mathbf{\delta} = (1,1,2,2,2,0,0)\\
   \mathbf{\beta} = (4,1,2,5,7,0,0)
  \end{array}$}
    {$\begin{array}{c} 
    \mathbf{\delta} = (1,1,2,2,2,2,0)\\
    \mathbf{\beta} = (4,1,2,5,7,3,0)
    \end{array}$}
    {$\begin{array}{c} 
    \mathbf{\delta} = (1,1,2,2,2,2,0)\\
    \mathbf{\beta} = (4,1,2,5,7,6,0)
    \end{array}$}
]
}
\end{center}
\caption{The node from equation~(\ref{expandExample}), 
written in node expansion format, is
expanded using the node expansion algorithm in
Table~\ref{tbl:nodeExpand}.}
\label{fig:nodeExpand}
\begin{center}
{\footnotesize
\qtreecenterfalse
\Tree
[.{$\begin{array}{c}
   \alpha_{d_1} = \langle a_4,a_1 \rangle\\
   \alpha_{d_2} = \langle a_2,a_5,a_7 \rangle
  \end{array}$}
    {$\begin{array}{c} 
    \alpha_{d_1} = \langle a_4,a_1 \rangle\\
    \alpha_{d_2} = \langle a_2,a_5,a_7,a_3 \rangle
    \end{array}$}
    {$\begin{array}{c} 
    \alpha_{d_1} = \langle a_4,a_1 \rangle\\
    \alpha_{d_2} = \langle a_2,a_5,a_7,a_3,a_6 \rangle
    \end{array}$}
]
}
\end{center}
\caption{The expansion in Figure~\ref{fig:nodeExpand} in
our original notation.}
\label{fig:nodeExpand2}
\end{figure}
Using this algorithm, we can grow the assignment tree
for any RDTA problem.  In Figure~\ref{fig:nodeExpand3} we show the
tree for the two vehicle two attacker example written in our
node expansion algorithm's notation.

\subsection{Search algorithm}
\label{sec:Branch}
To determine the order in which we expand nodes, we have tried several
tree search algorithms including the systematic search algorithms
breadth first search (BFS), depth first search (DFS)~\cite{Cormen}, and
A* search~\cite{stefik}. 
\begin{figure}
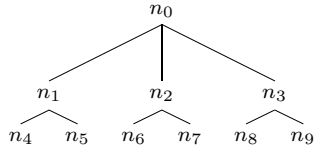

\begin{center}
{\footnotesize
\qtreecenterfalse
\Tree 
[.{$n_0$} 
  [.{$n_1$} 
    {$n_4$} 
    {$n_5$} 
  ] 
  [.{$n_2$} 
    {$n_6$} 
    {$n_7$} 
  ] 
  [.{$n_3$} 
    {$n_8$} 
    {$n_9$} 
  ] 
]
}
\end{center}
\caption{Example search tree used to illustrate the branching routine
in the branch and bound solver.
}
\label{fig:exampleTree}
\end{figure}
The A* search algorithm orders nodes according to a heuristic
branching function to help guide the search toward the optimal
assignment. We use the upper bound algorithm presented in 
Section~\ref{sec:ubAlgo} as the branching function. The lower bound
algorithm presented in Section~\ref{sec:lbAlgo} could also be used as
the branching function.

For example, consider a tree with three levels, where node $i$ is
labeled $n_i$ as shown in Figure~\ref{fig:exampleTree}. For this tree,
BFS gives the ordering
$(n_0,n_1,n_2,n_3,n_4,n_5,n_6,n_7,n_8,n_9)$,
and DFS gives the ordering
$(n_0,n_1,n_4,n_5,n_2,n_6,n_7,n_3,n_8,n_9)$.
Suppose the upper bound algorithm run at each node $i$ gives the
following  results: 
$J_{ub}(n_1)=3$,
$J_{ub}(n_2)=1$,
$J_{ub}(n_3)=2$,
$J_{ub}(n_4)=2$,
$J_{ub}(n_5)=1$,
$J_{ub}(n_6)=1$,
$J_{ub}(n_7)=1$,
$J_{ub}(n_8)=1$,
$J_{ub}(n_9)=0$.
A* BFS gives 
the ordering
$(n_0,n_2,n_3,n_1,n_6,n_7,n_9,n_8,n_5,n_4)$,
and A* DFS gives 
the ordering
$(n_0,n_2,n_6,n_7,n_3,n_9,n_8,n_1,n_5,n_4)$.

In A* search, the children of a node must be sorted with respect to
the branching function.  The maximum number of children that emanate
from any given node is the $nm$ children emanating from the root node.
Therefore, the maximum number of items that need to be sorted is $nm$.
To sort the children, we use Shell's method~\cite{Press92}, which runs
in $O((nm)^{3/2})$ time.  

\subsection{Upper bound algorithm}
\label{sec:ubAlgo}
In this section, we describe a fast algorithm that generates a
feasible complete assignment given any partial assignment. The cost of
the resulting complete assignment is an upper bound on the optimal
cost that can be achieved from the given partial assignment.  The idea
behind the upper bound algorithm is to assign unassigned attackers in
a greedy way.  At each step, we assign the attacker defender pair
that results in the minimum intercept time.  We proceed until all
attackers are assigned or until none of the remaining attackers can be
intercepted before entering the Defense Zone.  The details of this
algorithm, which runs in $O(nm^2)$ time, are listed in
Table~\ref{tbl:greedy}.

The input to the algorithm is a partial assignment given by an
intercept sequence $\alpha_d$ for each defender $d$ in $\mathcal{D}$
such that the set of unassigned attackers $\mathcal{A}_u$ is not
empty. In addition, we take as inputs the variables associated with
this partial assignment including the time for defender $d$ to
complete its intercept sequence $\alpha_d$, given by $t_d(m_d)$, and
binary variable $\gamma_a$ for each $a$ in the set of assigned
attackers $\mathcal{A}-\mathcal{A}_u$.

Given a partial assignment, the greedy step of the algorithm
determines the attacker in the set $\mathcal{A}_u$ that can be
intercepted in the minimum amount of time, denoted $a^*$.  The
corresponding defender that intercepts $a^*$ is denoted $d^*$. To
determine this defender, attacker pair $(d^*,a^*)$ we form a matrix
$C$ of intercept times. The matrix has size $|\mathcal{D}| \times
|\mathcal{A}_u|$, and its elements are given by
\begin{eqnarray}
c(d,a) := t_d(m_d) + \Delta t_{int}(d,a,t_d(m_d)),
\end{eqnarray}
for each $d$ in $\mathcal{D}$ and $a$ in $\mathcal{A}_u$. The element
$c(d,a)$ is the time it would take defender $d$ to complete its
intercept sequence $\alpha_d$ and then intercept attacker $a$. The
minimum of these times gives the desired defender, attacker pair
\begin{eqnarray}
c(d^*,a^*) = \min_{d\in\mathcal{D}, a\in \mathcal{A}_u} c(d,a).
\end{eqnarray}

If $c(d^*,a^*) = \infty$, no attacker can be intercepted before it
enters the Defense Zone. Thus, we set $\gamma_a := 1$ for each $a$ in
$\mathcal{A}_u$.  Then, we set $\mathcal{A}_u$ to the empty set
because all attackers are effectively assigned, and we use
equation~(\ref{eqn:objective}) to calculate the upper bound $J_{ub}$.
Otherwise, $c(d^*,a^*)$ is finite, and we add attacker $a^*$ to
defender $d^*$'s intercept sequence by incrementing $m_{d^*}$ by
one and setting $\alpha_{d^*}(m_{d^*}) := a^*$. Then, because $a^*$ has now
been assigned, we remove it from the set of unassigned attackers by
setting $\mathcal{A}_u := \mathcal{A}_u - \{ a^* \}$.
If $\mathcal{A}_u$ is not empty, we have a new partial assignment, and we
repeat the procedure. Otherwise, the assignment is
complete and we use equation~(\ref{eqn:objective}) to compute the upper bound
$J_{ub}$.

\begin{table}
\caption{Greedy upper bound algorithm}
\label{tbl:greedy}
\begin{center}
\framebox{\parbox{3.4in}{
\begin{algorithmic}[1]
\STATE Given a partial assignment: intercept sequence $\alpha_d$ for
each $d \in \mathcal{D}$, a nonempty set of unassigned attackers
$\mathcal{A}_u$, $t_d(m_d)$ for each $d \in \mathcal{D}$, and
$\gamma_a$ for each $a \in (\mathcal{A} - \mathcal{A}_u)$.
\STATE Initialize variables for unassigned attackers. 
Set $\gamma_a := 0$ for each $a$ in $\mathcal{A}_u$.
\STATE Calculate the elements of matrix $C$. For all $d \in
\mathcal{D}$ and $a \in \mathcal{A}_u$, set
\begin{eqnarray}
c(d,a) := t_d(m_d) + \Delta t_{int}(d,a,t_d(m_d)).
\nonumber
\end{eqnarray}
\WHILE {$\mathcal{A}_u$ not empty}
\STATE Find minimum element of $C$ given by 
\begin{eqnarray}
c(d^*, a^*) = \min_{d\in\mathcal{D},a\in\mathcal{A}_u} c(d,a).
\nonumber
\end{eqnarray}
\STATE If $c(d^*,a^*) = \infty$, no
attacker in the set $\mathcal{A}_u$ can be intercepted before
entering the Defense Zone.  Break out of the while loop.
\STATE Append attacker $a^*$ to defender $d^*$'s assignment
by setting $m_{d^*} := m_{d^*} + 1$ and
$\alpha_{d^*}(m_{d^*}) := a^*$. 
\STATE Update finishing time for $d^{*}$ by setting  
$t_{d^*}(m_{d^*}) := c(d^*,a^*)$.
\STATE Remove $a^*$ from consideration since it has been
assigned.  Set $c(d,a^*)$ to
$\infty$ for all $d \in \mathcal{D}$, and set $\mathcal{A}_u :=
\mathcal{A}_u - \{ a^* \}$.
\STATE Update matrix for defender $d^*$. For all attackers $a \in
\mathcal{A}_u$, set 
\begin{eqnarray}
c(d^*,a) := t_{d^*}(m_{d^*}) + \Delta t_{int}
(d^*,a,t_{d^*}(m_{d^*})).
\nonumber
\end{eqnarray} 
\ENDWHILE
\STATE For each $a$ in $\mathcal{A}_u$, set $\gamma_a := 1$.
\STATE Set 
\begin{eqnarray}
J_{ub} := \sum_{a \in \mathcal{A}} \gamma_a + \epsilon \max_{d \in
\mathcal{D}} t_d(m_d).
\nonumber 
\end{eqnarray}
\end{algorithmic}
}}
\end{center}
\end{table}

\subsection{Lower bound algorithm}
\label{sec:lbAlgo}
Here we describe a fast algorithm that generates a lower bound on the
cost that can be achieved from any given partial assignment.  The idea
behind the algorithm is to use the principle of simultaneity.  In
assigning attackers from $\mathcal{A}_u$, we assume each defender can
pursue multiple attackers simultaneously. The result is a potentially
infeasible complete assignment because simultaneity is physically
impossible.  Because no feasible assignment can do better, the cost of
this assignment is a lower bound on the optimal cost that can be
achieved from the given partial assignment.  The algorithm, which runs
in $O(nm)$ time, is listed in Table~\ref{tbl:LB}.

Similar to the upper bound algorithm, the input to the lower bound
algorithm is a partial assignment. This includes an intercept sequence
$\alpha_d$ for each defender $d$ in $\mathcal{D}$ with $\mathcal{A}_u$
nonempty, $t_d(m_d)$ for each defender $d$ in $\mathcal{D}$, and
$\gamma_a$ for each attacker $a$ in $\mathcal{A}-\mathcal{A}_u$.

Each attacker $a$ in $\mathcal{A}_u$ is assigned a defender as
follows: Form a matrix $C$ with elements
\begin{eqnarray}
c(d,a) := t_d(m_d) + \Delta t_{int}(d,a,t_d(m_d)),
\end{eqnarray}
for all $d$ in $\mathcal{D}$ and $a$ in $\mathcal{A}_u$. Element
$c(d,a)$ is equal to the time it takes $d$ to intercept the attackers in its
intercept sequence $\alpha_d$ plus the time it would take to
subsequently intercept attacker $a$.
For each $a$ in $\mathcal{A}_u$, find the defender, denoted $d^*$,
that can intercept $a$ in minimal time
\begin{eqnarray}
  c(d^*,a) = \min_{d\in \mathcal{D}} c(d,a).
\end{eqnarray}

If $c(d^*,a) = \infty$, we set $\gamma_a := 1$ because no defender can
intercept attacker $a$ before it enters the Defense Zone.  Otherwise,
we set $\gamma_a := 0$ because defender $d^*$ can intercept attacker
$a$ before it enters the Defense Zone.  The lower bound is therefore
give by
\begin{eqnarray}
J_{lb} := \sum_{a \in \mathcal{A}} \gamma_a + 
\epsilon \max_{
\{a \in \mathcal{A} : \gamma_a=0 \} }
\left(
\min_{d \in \mathcal{D}} c(d,a)
\right).
\end{eqnarray}

\begin{table}
\caption{Lower bound algorithm}
\label{tbl:LB}
\begin{center}
\framebox{\parbox{3.4in}{
\begin{algorithmic}[1]
\STATE Given a partial assignment: intercept sequence $\alpha_d$ for
each $d \in \mathcal{D}$, a nonempty set of unassigned attackers
$\mathcal{A}_u$, $t_d(m_d)$ for each $d \in \mathcal{D}$, and
$\gamma_a$ for each $a \in (\mathcal{A} - \mathcal{A}_u)$.
\STATE Calculate the elements of matrix $C$. For all $d \in
\mathcal{D}$ and $a \in \mathcal{A}_u$, set
\begin{eqnarray}
c(d,a) := t_d(m_d) + \Delta t_{int}(d,a,t_d(m_d)).
\nonumber
\end{eqnarray}
\FORALL {$a \in \mathcal{A}_u$}
\STATE Find minimum element of $a$th column of $C$ given by
\begin{eqnarray}
  c(d^*,a) = \min_{d\in \mathcal{D}} c(d,a).\nonumber
\end{eqnarray}
\STATE {\bf if} $c(d^*,a) = \infty$ {\bf then} 
set $\gamma_a := 1$.
\STATE {\bf else} set $\gamma_a := 0$.
\ENDFOR
\STATE Set 
\begin{eqnarray}
J_{lb} := \sum_{a \in \mathcal{A}} \gamma_a + 
\epsilon \max_{
\{a \in \mathcal{A} : \gamma_a=0 \} }
\left(
\min_{d \in \mathcal{D}} c(d,a)
\right).
\nonumber
\end{eqnarray}
\end{algorithmic}
}}
\end{center}
\end{table}

\section{Analysis of the solver}
\label{sec:analysis}
In this section, we explore the average case computational complexity
of the branch and bound algorithm by solving randomly generated
instances.  Each instance is generated by randomly selecting
parameters from a uniform distribution over the intervals defined
below.  The computations were performed on a PC with Intel PIII 550MHz
processor, 1024KB cache, 3.8GB RAM, and Linux.  For all instances
solved, processor speed was the limiting factor, not memory.

\subsection{Generating random instances}
\label{sec:gen}
The initial position of each attacker is taken to be in an annulus
centered on the playing field. The radius of the initial position,
denoted $r_a$, is chosen at random from a uniform distribution over
the interval $[r_a^{\min},r_a^{\max}]$. The angle of the initial
position, denoted $\theta_a$, is chosen from a uniform distribution
over the interval $(0,2\pi]$ (all other angles used in this section
$\phi_{a}$, $\theta_d$, and $\phi_{d}$ are also chosen
from a uniform distribution over the interval $(0,2\pi]$).  The
magnitude of attacker $a$'s velocity, denoted $v_a$, is chosen at
random from a uniform distribution over the interval
$[v_a^{\min},v_a^{\max}]$.  The initial state of the attacker
is given by 
\begin{eqnarray} 
&&p(0) = r_a \cos(\theta_a),\mbox{ }q(0) = r_a \sin(\theta_a)\nonumber\\
&&\dot{p} = v_a \cos(\phi_a),\mbox{ }\dot{q} = v_a \sin(\phi_a).  
\end{eqnarray}

The initial position of each defender is taken to be in a smaller
annulus, also centered on the playing field. The radius of the initial
position, denoted $r_d$, is chosen at random from a uniform
distribution over the interval $[r_d^{\min},r_d^{\max}]$.  The
magnitude of defender $d$'s velocity, denoted $v_d$, is chosen at
random from a uniform distribution over the interval
$[v_d^{\min},v_d^{\max}]$.  The initial state of the defender is
given by
\begin{eqnarray}
&&x(0) = r_d \cos(\theta_d),\mbox{ }y(0) = r_d \sin(\theta_d)\nonumber\\
&&\dot{x}(0) = v_d \cos(\phi_d),\mbox{ }
\dot{y}(0) = v_d \sin(\phi_d). 
\end{eqnarray}

For the instances generated in this paper, we set $R_{dz}=2.0$ and
take the parameters from the following intervals: $r_a \in [7.5,15.0]$,
$v_d = 1.0$, $r_d \in [\sqrt{2}R_{dz},2\sqrt{2}R_{dz}]$, and $v_d \in
[0.5,1.0]$. In Section~\ref{sec:phaseTransitions}, we study
the RDTA problem with variations in the velocity parameters
$v_a$ and $v_d^{\max}$.

\subsection{Average case computational complexity}
\label{sec:Performance}
\begin{figure}
\centering
\includegraphics[width=200pt]{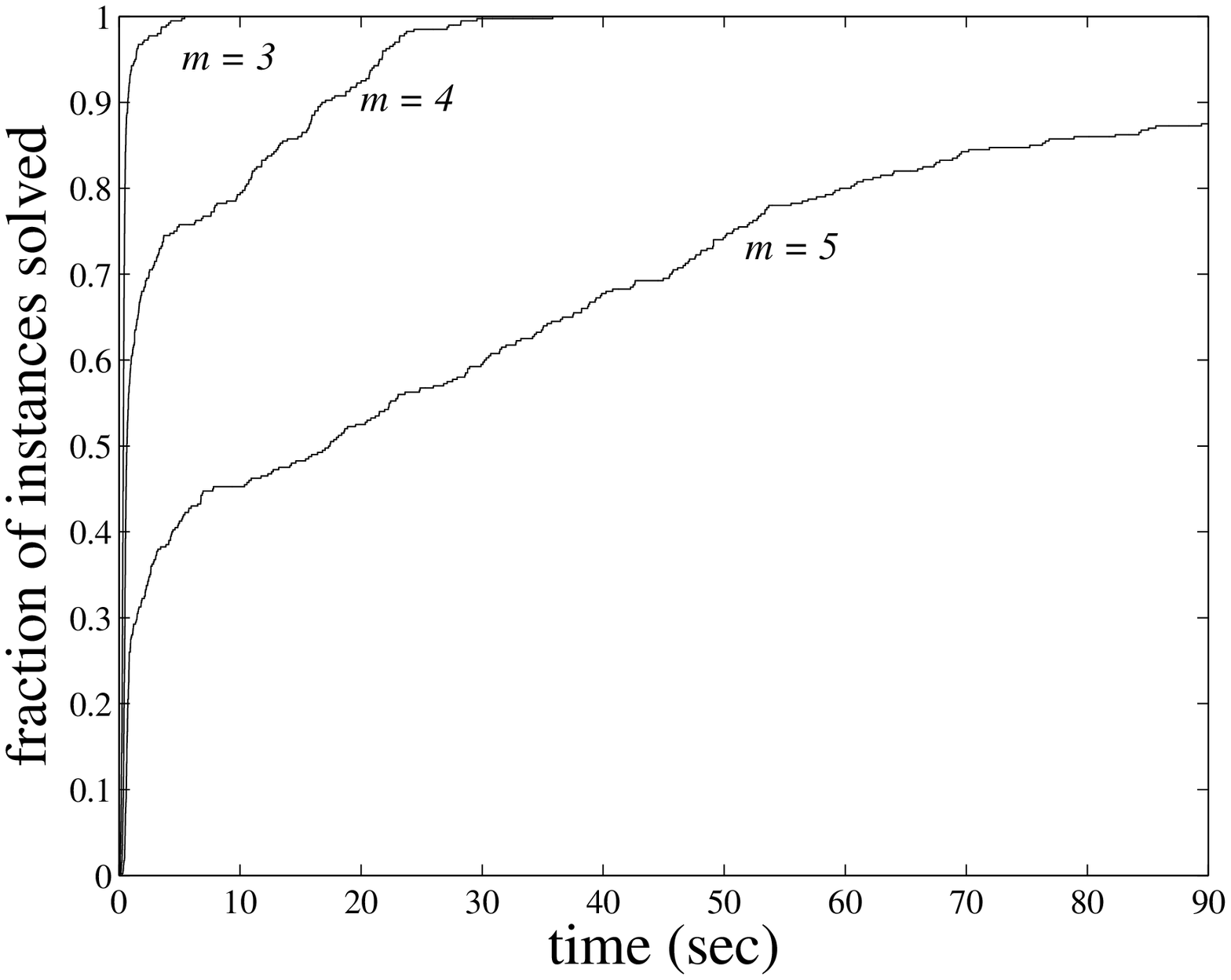}
\includegraphics[width=200pt]{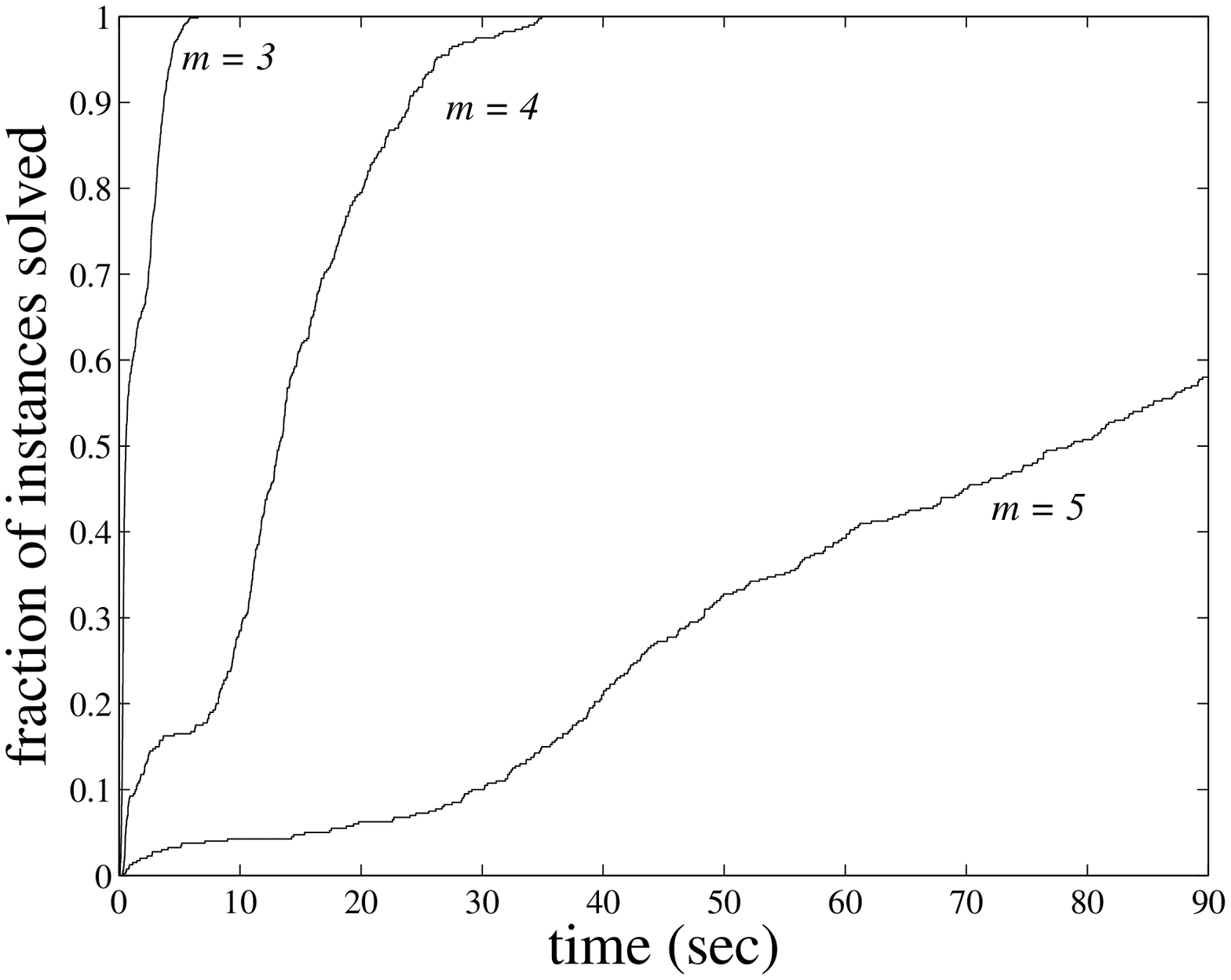}
\caption{
The fraction of instances solved versus computation time for the
branch and bound solver. On top, the cost is the number of attackers
that enter the Defense Zone ($\epsilon = 0$ in
equation~(\ref{eqn:objective})), and on bottom, the cost includes a
secondary component ($\epsilon = 0.01$ in
equation~(\ref{eqn:objective})). For each curve, 400 random instances
of the RDTA problem
were solved. The values of the parameters are $n=3$ and $m = 3,4,5$.  
}
\label{fig:ctime1}
\end{figure}

In this section, we present the results of an average case
computational complexity study on the branch and bound solver.
A particular
problem instance is considered solved when the strategy that minimizes
the cost is found.
In
Figure~\ref{fig:ctime1}, we plot the fraction of instances solved
versus computation time. In the figure on top, the cost function is
the number of attackers that enter the Defense Zone ($\epsilon=0$ in
equation~(\ref{eqn:objective})).  Solving these instances becomes
computationally intensive for modest size problems. For example, when
$n=3$ and $m=5$, 80\% of the instances are solved in 60 seconds
or less. In the figure on bottom, in addition to the primary component
of the cost function, the cost function includes a secondary component
($\epsilon = 0.01$ in equation~(\ref{eqn:objective})).  The secondary
component is the time it takes to intercept all attackers that can be
intercepted. Solving these instances of the problem is more
computationally intensive than the $\epsilon = 0$ case. For example,
when $n=3$ and $m=5$, only 40\% of the problems are solved in 60
seconds or less.

The increase in average case computational complexity for the
$\epsilon > 0$ case is expected because the cost function has an
additional component to be minimized, which is independent of the
primary component. In a case where the primary component is at a
minimum, the algorithm will proceed until it proves that the
combination of primary and secondary components is minimized.

If it is given enough time, the branch and bound solver finds the
optimal assignment, but the average case computational complexity is
high. Therefore, using the algorithm to solve for the optimal
assignment in real-time is infeasible for most applications.  However,
the best assignment found in the allotted time window for planning
could be used in place of the optimal assignment. In this case, it is
desirable that the algorithm converge to a near-optimal solution
quickly.  

To learn more about the convergence rate of the branch and bound
solver, we look at the rate at which the best upper bound
$J_{ub}^{best}$ decreases with branches taken in the search tree.
Because the branch and bound algorithm is an exact method,
$J_{ub}^{best}$ eventually converges to $J_{opt}$. We define the
percent difference from optimal as follows: Let
$J_{opt}^{(i)}$ be the optimal cost for instance $i$. Let
$J_{ub}^{(i)}(k)$ be the best upper bound found after $k$ branches for
instance $i$. Let $\hat{J}_{opt}$ be the mean of the set
$\{J_{opt}^{(i)}: i=1,\ldots,N \}$, and let $\hat{J}_{ub}(k)$ be the
mean of the set $\{J_{ub}^{(i)}(k): i=1,\ldots,N \}$, where $N$ is the
number of instances. The percent
difference from optimal is given by  
\begin{eqnarray} 
\mbox{PD}(k) = 100 \frac{\hat{J}_{ub}(k) - \hat{J}_{opt}}{\hat{J}_{opt}}.
\end{eqnarray} 

\begin{figure}
\centering
\includegraphics[width=220pt]{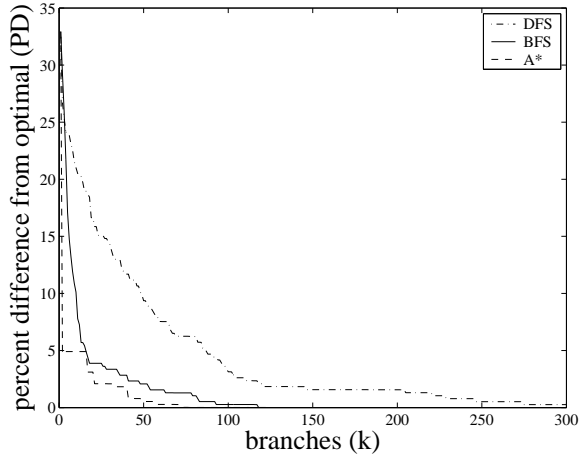}
\caption{The average convergence rate for the branch and bound solver
using each of the three branching routines BFS, DFS, and A* search. We
plot the percent difference from optimal $\mbox{PD}(k)$ 
versus the number
of branches $k$ explored.  For each curve 400 random instances of
RDTA were solved.  The parameters values are $\epsilon = 0.01$, 
$n=3$, and $m=5$.
}
\label{fig:converg1}
\end{figure}

In Figure~\ref{fig:converg1}, we plot PD$(k)$ versus the number of
branches $(k)$ for instances involving three defenders ($n=3$) and
five attackers ($m=5$).  At the root node ($k=1$), the greedy
algorithm is applied.  Exploration of the tree does not occur at this
point.  Therefore, the three branching routines produce the same
result, $\mbox{PD}(1) = 33\%$.  This means that $\hat{J}_{ub}(1) -
\hat{J}_{opt} = 0.33\hat{J}_{opt}$, or $\hat{J}_{ub}(1) = 1.33
\hat{J}_{opt}$.  In other words, the average cost of the assignment
generated by the greedy algorithm is 1.33 times the average optimal
cost.  At one branch into the tree ($k = 2$), both DFS and BFS
generate assignments with $\mbox{PD}(2) = 28\%$, and the A* search
generates assignments with $\mbox{PD}(2) = 5\%$. Therefore, after
only two steps, the branch and bound algorithm using A* search
generates an assignment that, on average, has cost only 1.05 times the
cost of the optimal assignment. 

For the instances solved here, the branch and bound solver with A*
search converges to the optimal assignment in an average of 8 
branches, and it takes an average of 740 branches to prove that the
assignment is optimal. Therefore, the solver converges to the optimal
solution quickly, and the computational complexity that we observed
(Figure~\ref{fig:ctime1}) is due to the time needed to prove
optimality.

These results are encouraging for real-time implementation of the
algorithm.  The results show that a very good assignment is generated
after a short number of branches.  There is a trade-off between
optimality and computation time that can be tuned by deciding how deep
into the tree to explore.  Going deeper into the tree will generate
assignments that are closer to optimal, but at the same time, results
in an increased computational burden.  The parameter to be tuned is
the maximum number of branches to allow the search procedure to
explore, denoted ${\tt kMax}$.

\begin{figure}
\centering
\includegraphics[width=220pt]{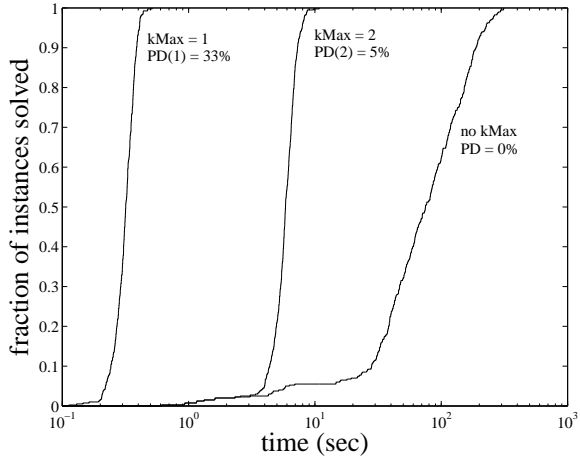}
\caption{The fraction of instances solved versus computation time for
the branch and bound solver (using A*) with ${\tt kMax}= 1,2,\mbox{ and
}\infty$.  The ${\tt kMax}$ variable controls the maximum number of branches
explored. We vary it from ${\tt kMax} = 1$, which is a greedy search, to
${\tt kMax} = \infty$, which is exhaustive search.  For each curve, 400
random instances of the RDTA problem was solved.  For these problems
the parameter values are $\epsilon = 0.01$, $n=3$, and $m = 5$.
}
\label{fig:ctimeComp}
\end{figure}

To study the computational complexity as ${\tt kMax}$ is tuned, we
look at versions of the algorithm (using A*) with ${\tt kMax}=1$ (greedy
algorithm), ${\tt kMax}=2$, and ${\tt kMax}=\infty$ (exact algorithm).  These
three cases generate assignments with average percent difference from
optimal given by PD(1)=33\%, PD(2)=5\%, and PD($\infty$)=0\%
respectively. The results are shown in Figure~\ref{fig:ctimeComp}.
The algorithm with ${\tt kMax}=2$ gives a good balance between optimality
and computation time.

\subsection{Phase Transitions}
\label{sec:phaseTransitions}
The RDTA problem is NP-hard~\cite{Garey}, which can be shown by
reduction using the traveling salesman problem.
This is a worst case result that says nothing about the
average case complexity of the algorithm or the complexity with
parameter variations.  In this section, we study the complexity of the
RDTA problem as parameters are varied. We perform this study
on the decision version of the problem.

\emph{RoboFlag Drill Decision Problem (RDD)}:
Given a set of defenders $\mathcal{D}$ and a set of attackers
$\mathcal{A}$, is there a complete assignment such that no attacker
enters the Defense Zone?

First, we consider variations in the ratio of attacker velocity to
maximum defender velocity, denoted $vA/vD$ in this section.  When the
ratio is small, the defenders are much faster than the attackers.  It
should be easy to quickly find an assignment such that all attackers
are intercepted.  When the ratio is large, the attackers are much
faster than the defenders. In this case, it is difficult for the
defenders to intercept all of the attackers, which should be easy to
determine. 

The interesting question is whether there is a transition from being
able to intercept all the attackers (all \emph{yes} answers to the RDD
problem) to not being able to intercept all attackers (all \emph{no}
answers to the RDD problem). Is this transition sharp? Are there
values of the ratio for which solving the RDD is difficult?

\begin{figure}
\centering
\includegraphics[width=220pt]{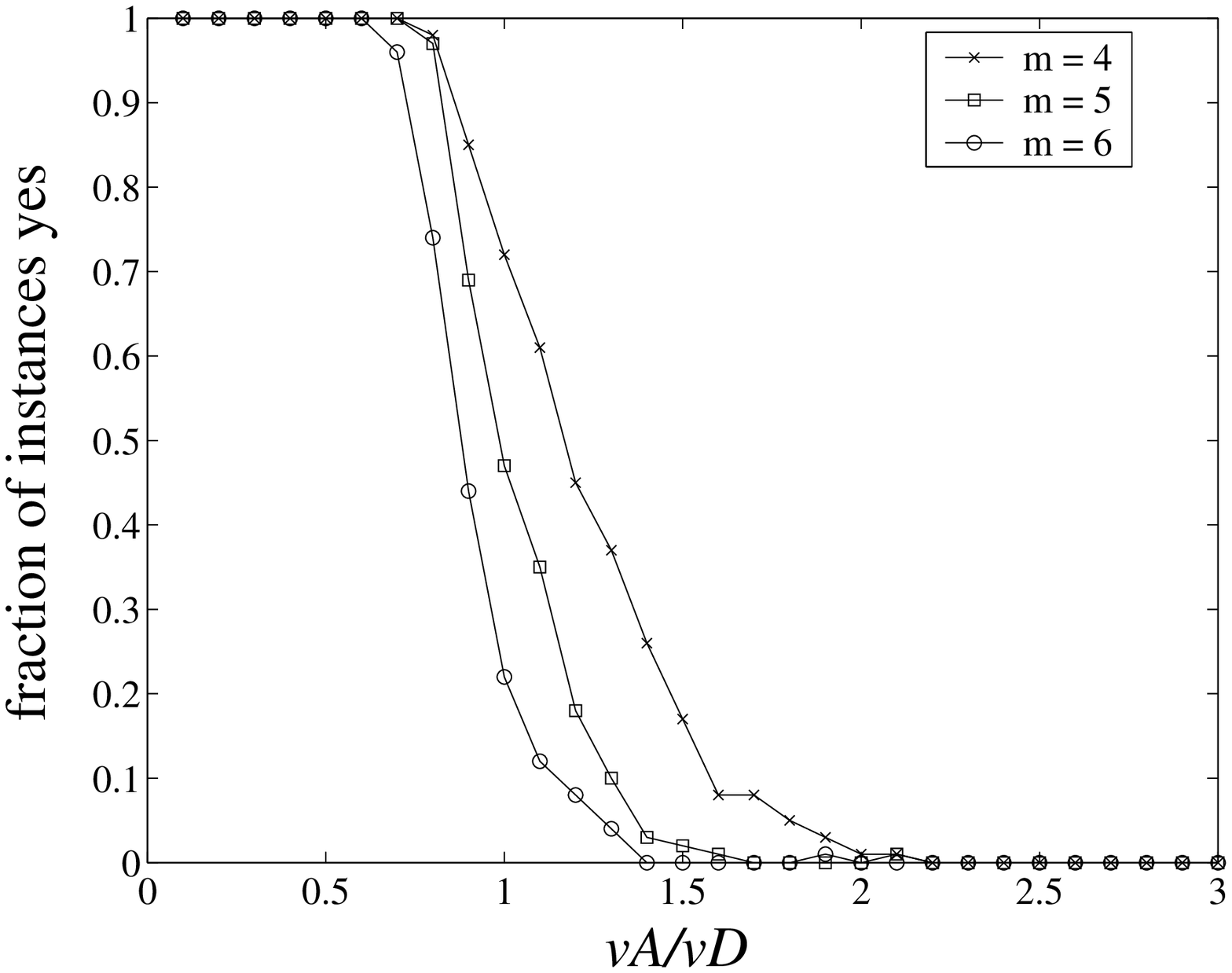}
\includegraphics[width=220pt]{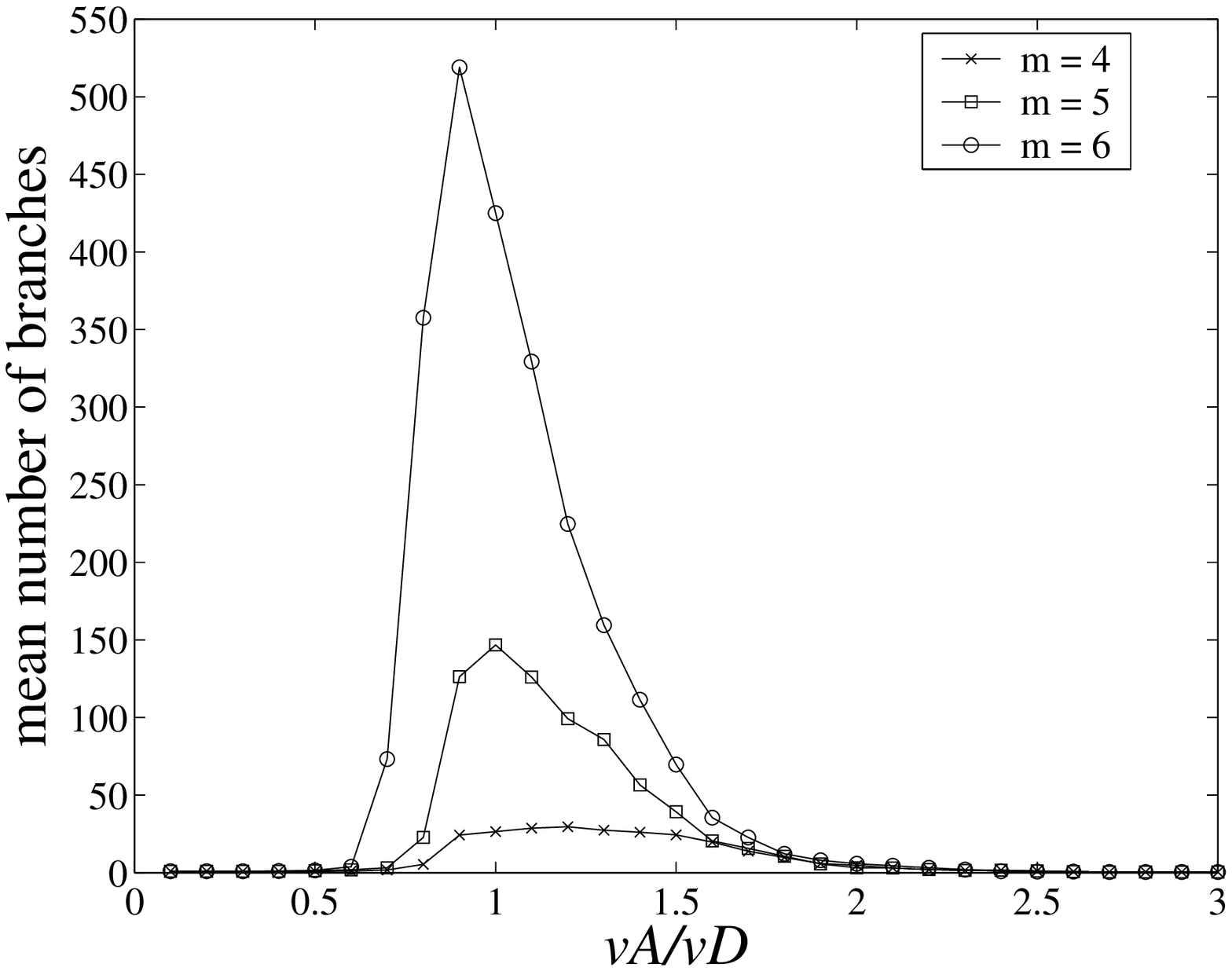}
\caption{The phase transition of the RDD problem in the ratio of
attacker velocity to maximum defender velocity ($vA/vD$). The figure
on top shows the fraction of instances that evaluate to \emph{yes}
versus the velocity ratio. The figure on bottom shows the mean number
of branches needed to solve the problem versus the velocity ratio.
The phase transition occurs at a velocity ration of approximately 1.
For each curve, 100 random instances of the RDD problem were solved.
In these figures, $n=3$.
}
\label{fig:pt1}
\end{figure}

For each value of the velocity ratio, we generated random instances of
the RDD problem and solved them with the branch and bound solver.  The
results are shown in Figure~\ref{fig:pt1}. The figure on top shows the
fraction of instances that evaluate to \emph{yes} versus the velocity
ratio. The figure on bottom shows the mean number of branches required
to solve an instance versus the velocity ratio.  There is a sharp
transition from all instances \emph{yes} to all instances \emph{no}.
This transition occurs approximately at $vA/vD = 1$ for the $n=3$,
$m=5$ case. At this value of the ratio, there is a spike in
computational complexity.  This easy-hard-easy behavior is indicative
of a phase transition~\cite{Bejar,Monasson}.

\begin{figure}
\centering
\includegraphics[width=220pt]{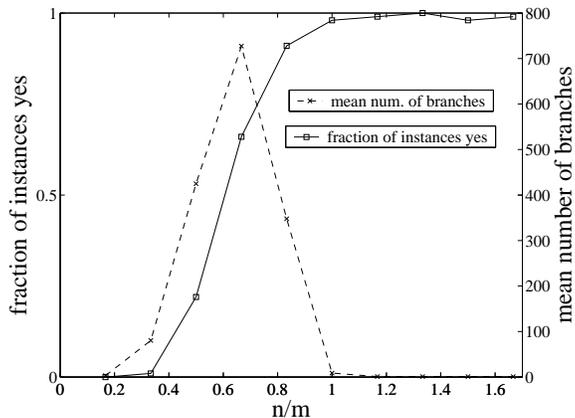}
\caption{The phase transition of the RDD
problem in the ratio of defenders to attackers ($n/m$).
The solid line shows the fraction of instances that evaluate to
\emph{yes} versus the ratio.  The dashed line shows the mean number of
branches needed to solve the problem versus the ratio. 
For each curve, 100 random instances of the RDD problem were solved.
The velocities are $vD = vA = 1$.
}
\label{fig:pt2}
\end{figure}

We also study the RDD problem with variations in the ratio of
defenders to attackers, denoted $n/m$, with $vD=vA=1$.  For small
values of $n/m$, the number of attackers is much larger than the
number of defenders, and it should be easy to determine that the team
of defenders cannot intercept all of the attackers. In this case, most
instances should evaluate to \emph{no}.  For large values of $n/m$,
the number of defenders is much larger than the number of attackers,
and it should be easy to find an assignment in which all attackers are
denied from the Defense Zone. In this case, most instances should
evaluate to \emph{yes}. The results are shown in Figure~\ref{fig:pt2},
where it is clear that our expectations proved correct.  In between
the extremes of the $n/m$ ratio, there is a phase transition at a
ratio of approximately $n/m = 0.65$.

In general, these experiments show that when one side dominates the
other (in terms of the number of vehicles or in terms of the
capabilities of the vehicles) the RDD problem is easy to solve. When
the capabilities are comparable (similar numbers of vehicles, similar
performance of the vehicles), the RDD is much harder to solve.  This
behavior is similar to the complexity of balanced games like
chess~\cite{Herik}. In Section~\ref{discussion}, we discuss how
knowledge of the phase transition can be exploited to reduce
computational complexity.

\section{Multi-level implementation} \label{sec:mpc}
Now that we have a fast solver that generates near-optimal assignments,
we test it in a dynamically changing environment. We consider the
RoboFlag Drill problem with attackers that have a simple
noncooperative strategy built in, which is unknown to the defenders. The
hope is that frequent replanning, at all levels of the hierarchical
decomposition, will mitigate our assumption that the attackers move
with constant velocity.

\begin{figure}
\centering
\includegraphics[width=150pt]{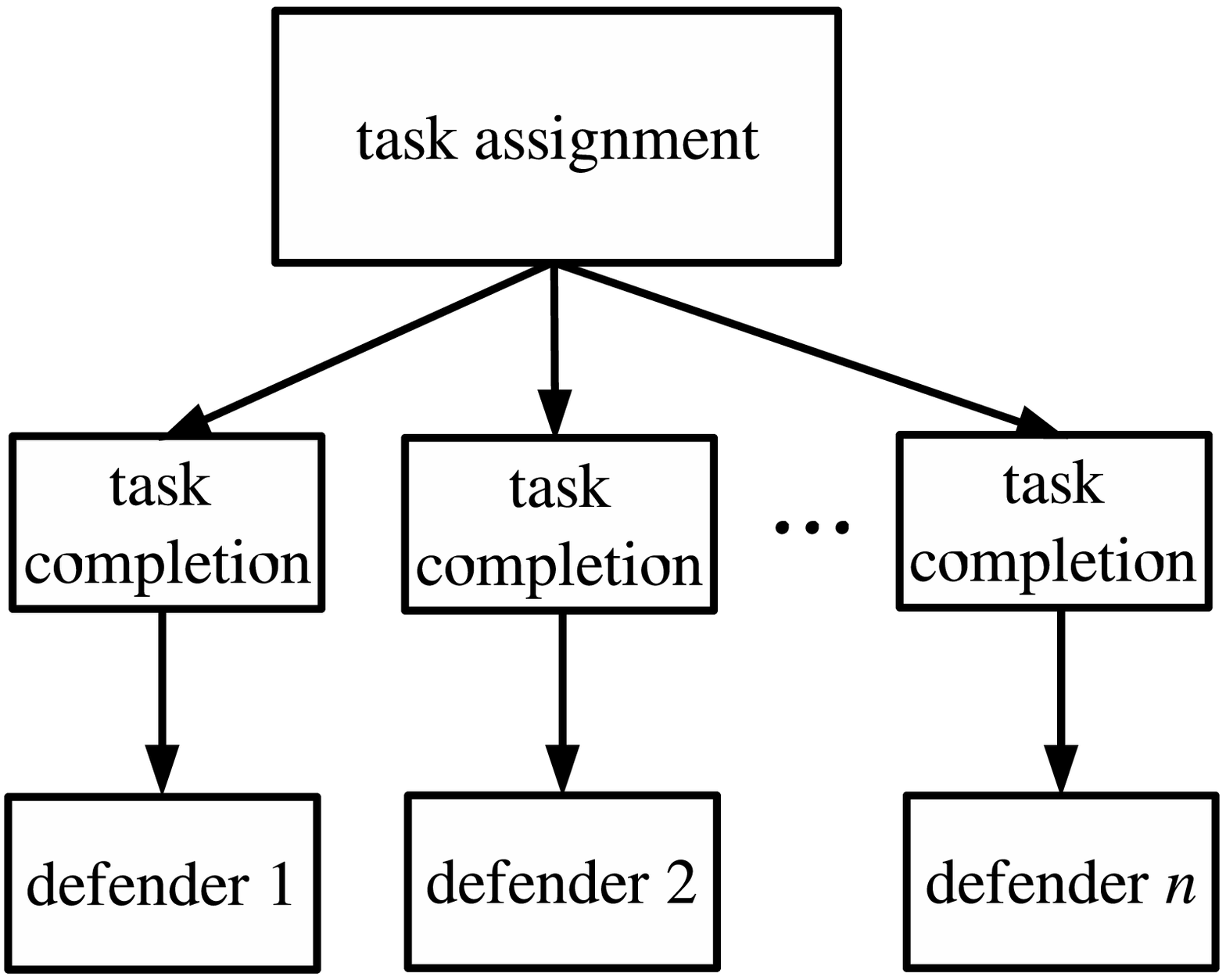}
\put(10,102){$R_{TA}$}
\put(10,55){$R_{TC}$}
\caption{The multi-level architecture for the defending vehicles used
in our implementation.}
\label{fig:vhierarchy}
\end{figure}

We use a multi-level receding horizon architecture, shown in
Figure~\ref{fig:vhierarchy}, to generate the defenders' strategy.  The
task assignment module at the top level implements the branch and
bound algorithm presented in this paper. It generates the assignment
$\alpha_d$ for each defender $d$, sending new assignments to the
middle level of the hierarchy at the rate $R_{TA}$.  Therefore, the
algorithm returns the best assignment computed in the time window
$1/R_{TA}$.  

There is a task completion module for each defender at the middle
level of the hierarchy, which receives an updated assignment
$\alpha_d$ from the task assignment module at the rate $R_{TA}$. At a
rate $R_{TC}$, the task completion module generates a trajectory from
defender $d$'s current state to a point that will intercept attacker
$\alpha_d(1)$ assuming the attacker moves at constant velocity.  If
attacker $\alpha_d(1)$ is intercepted, a trajectory to intercept
attacker $\alpha_d(2)$ is generated, and so on.  

The vehicle module at the bottom of the hierarchy receives an updated
trajectory from the task completion module at the rate $R_{TC}$.  The
module propels the vehicle along this trajectory until it receives an
update.

\begin{figure}
\centering
\includegraphics[width=150pt]{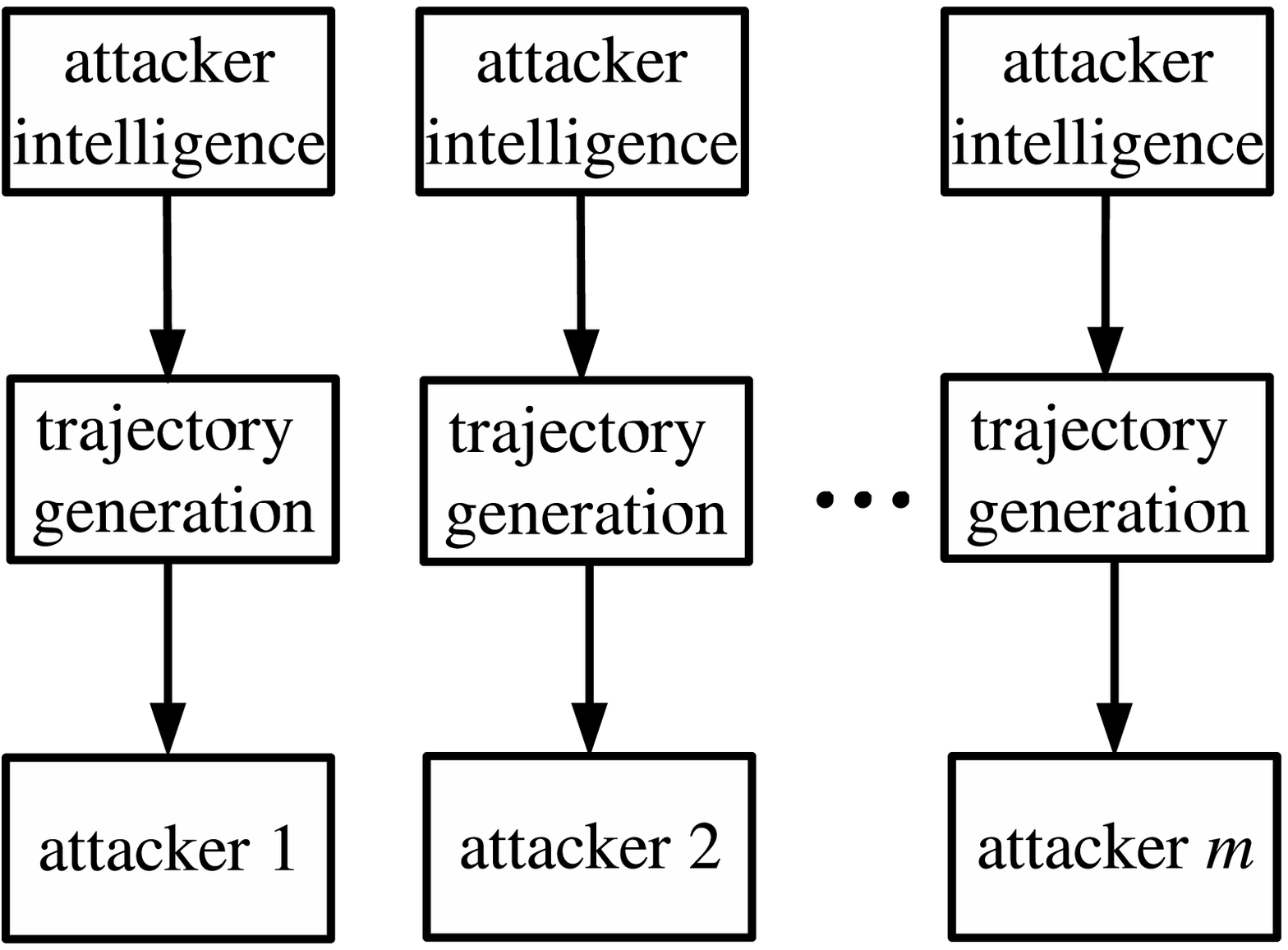}
\put(12,94){$R_{I}$}
\put(12,55){$R_{TG}$}
\caption{The multi-vehicle architecture for the attackers
used to test the defender architecture. }
\label{fig:ahierarchy}
\end{figure}

The attackers are taken to be the same vehicles as the defenders
(described in Appendix~\ref{sec:vehicleDynamics}).  For the attacker
intelligence, we use the architecture shown in
Figure~\ref{fig:ahierarchy}.  The levels of the hierarchy are
decoupled, so each attacker acts independently.  The simple
intelligence for each attacker is contained in the top level of the
hierarchy.  The primary objective is to arrive at the origin of the
field in minimum time.  However, the attacker tries to avoid the
defenders if they get too close. The radius of each defender is
artificially enlarged by a factor $\beta>1$. If the artificially
enlarged defenders obstruct an attacker's path toward the origin, the
attacker treats them as obstacles, finding a destination that results
in an obstacle free path. The destination is found using a simple
reactive obstacle avoidance routine used in
RoboCup~\cite{d'andrea01,Stone01}.  The attacker intelligence module
runs at the rate $R_{I}$. 

The trajectory generation module at the middle of the hierarchy
receives an updated destination at the rate $R_I$. The module
generates a trajectory from the current state of the attacker to the
destination with zero final velocity at the rate $R_{TG}$, using
techniques from~\cite{Nagy04}.  The vehicle module at the bottom level
of the hierarchy is the same as that for the defenders.  

Because the algorithms are more computationally intensive for the
higher levels of the hierarchy than the lower levels, the rates are
constrained as follows: $R_{TA} < R_{TC}$ and $R_{I} < R_{TG}$. In
the simulations that follow, we take $R_{TC}=R_{TG}$ because the
middle levels of the two hierarchies are comparative computationally.
We also set $R_I=R_{TG}/10$. Therefore, if the trajectory generation
module replans every time unit, the attacker intelligence module
replans every ten time units. 

First, note that when both $R_{TA}$ and $R_{TC}$ are zero, there is no
replanning. In this case, all attackers usually enter the Defense
Zone. They easily avoid the defenders because the defenders execute a
fixed plan, which becomes obsolete once the attackers start using their
intelligence.

Next, we present simulation results of the RoboFlag Drill with
intelligent attackers and defenders. We consider problems with eight
defenders ($n=8$), four attackers ($n=4$), $vA=vD$, $R_{TC}=R_{TG}>0$,
and $R_I=R_{TG}/10$. We consider several different values of the rate
at which the task assignment module replans ($R_{TA}$). For each
value, we solve 200 randomly generated instances of the problem.  As
an evaluation metric, we use the average number of attackers that
enter the Defense Zone during play.

For the case $R_{TA}=0$, there is no replanning at the task assignment
level. Replanning only occurs at the task completion level.  The
defenders are given a plan from the task assignment module at the
beginning of play. Each defender executes its assignment throughout,
periodically recalculating the trajectory it must follow to intercept
the next attacker in its sequence. For this case, on average, 58\% of
the attackers enter the Defense Zone during play.

For the case $R_{TA}>0$, replanning occurs at both the task assignment
level and the task completion level of the hierarchy. In addition to
recomputing trajectories to intercept the next attacker in each
defender's assignment, the defender assignments are recomputed. This
redistributes tasks based on the current state of the dynamically
changing environment, providing feedback.  For $R_{TA} = R_{TC}/40$,
$R_{TC}/20$, and $R_{TC}/15$, an average of 38\%, 34\%, and 32.5\% of
the attackers enter the Defense Zone during play, respectively.
Therefore, replanning at the task assignment level has helped increase
the utility of the strategies generated for the team of defenders.

\begin{figure}
\centering
\includegraphics[width=250pt]{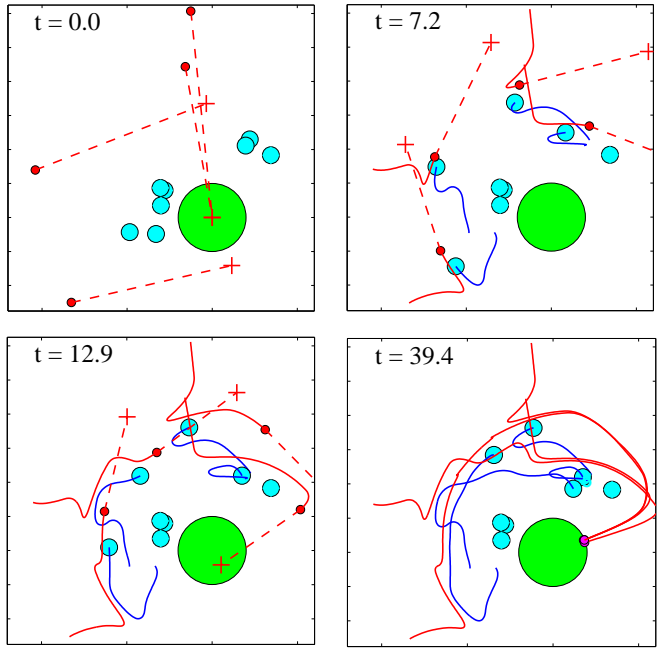}
\caption{
Snapshots of the RoboFlag Drill simulation with defender
replanning at the task completion level only ($R_{TA}=0$).
In this case, all attackers enter the Defense Zone.
The large
circles are the defenders, and the small circles are the attackers. The
solid lines are trajectories. Each cross connected to a dashed line is
an attacker's desired destination.
}
\label{rd_instance_noreplan}
\end{figure}

\begin{figure}
\centering
\includegraphics[width=250pt]{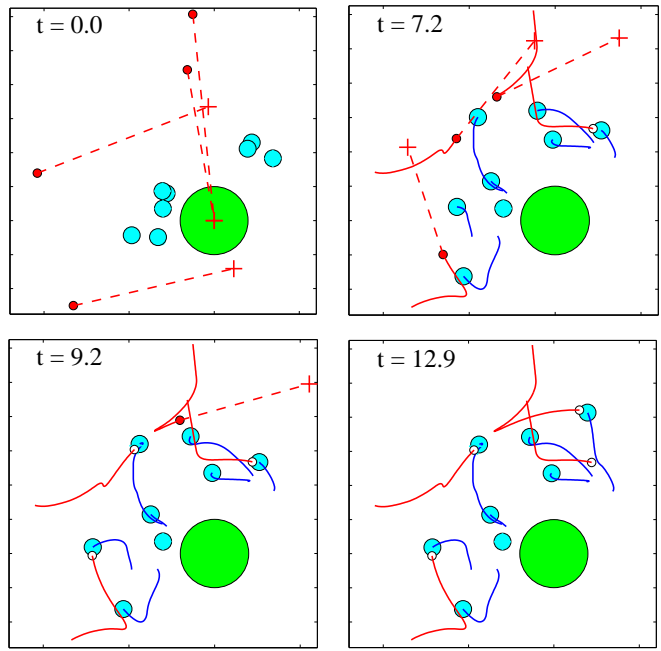}
\caption{
Snapshots of the RoboFlag Drill simulation with defender
replanning at the task completion level and task assignment
level ($R_{TA} = R_{TC}/15$).  Because there is replanning at both
levels, the defenders cooperate to intercept all attackers. The large
circles are the defenders, and the small circles are the attackers. The
solid lines are trajectories. Each cross connected to a dashed line is
an attacker's desired destination.
}
\label{rd_instance_replan}
\end{figure}

In Figure~\ref{rd_instance_noreplan}, we show snapshots of an instance
of the RoboFlag Drill simulation for the case where the defenders do
not replan at the task assignment level. In this case, all attackers
enter the Defense Zone.  In Figure~\ref{rd_instance_replan}, we show
snapshots of the same instance of the RoboFlag Drill simulation, but
in this case, the defenders replan at the task assignment level
($R_{TA} = R_{TC}/15$).  The defenders cooperate to deny
all attackers from the Defense Zone. For example, the two defenders at
the lower left of the field cooperate to intercept an
attacker.

\section{Discussion}
\label{discussion}

We developed a decomposition approach that generates cooperative
strategies for multi-vehicle control problems, and we motivated the
approach using an adversarial game called RoboFlag.  In the game, we
fixed the strategy for one team and used our approach to generate
strategies for the other team. By introducing a set of tasks to be
completed by the team and a task completion method for each vehicle,
we decomposed the problem into a high level task assignment problem
and a low level task completion problem.  We presented a branch and
bound solver for task assignment, which uses upper and lower bounds on
the optimal assignment to prune the search space. The upper bound
algorithm is a greedy algorithm that generates feasible assignments.
The best greedy assignment is stored in memory during the search, so
the algorithm can be stopped at any point in the search and a feasible
assignment is available. 

In our computational complexity study, we found that solving the task
assignment problem is computationally intensive, which was expected
because the problem is NP-hard.  However, we showed that the solver
converges to the optimal assignment quickly, and takes much more time
to prove the assignment is optimal. Therefore, the solver can be run
in a time window to generate near-optimal assignments for real-time
multi-vehicle strategy generation. To increase the speed of the
algorithm, it may be advantageous to distribute the computation over
the set of vehicles~\cite{Ralphs03}, taking advantage of the
distributed structure of the problem.

We also studied the computational complexity of the solver as
parameters were varied. We varied the ratio of the maximum velocities
of the opposing vehicles, and we varied the ratio of the number of
vehicles per team. We found that when one team has a capability
advantage over the other, such as a higher maximum velocity or more
vehicles, the solution to the task assignment problem is easy to
generate. However, when the teams are comparable in capability,
finding the optimal assignment to the problem is much more
computationally intensive. This type of analysis can help in deciding
how many vehicles to deploy in an adversarial game and what
capabilities the vehicles should have. In addition, knowledge of the
phase transition may be exploited to reduce computational complexity.
In~\cite{schneider96,schneider03}, phase transition `backbones' are
exploited to decompose combinatorial problems into many separate
subproblems, which are much less computationally intensive. This
decomposition is amenable to parallel computation. In~\cite{Gomes00},
it is shown that the hardness of a problem depends on the parameters
of the problem (as we showed above) and the details of the algorithm
used to solve the problem. Therefore, it is possible that the hard
instances of our problem, which lie along the phase transition, may be
solved faster if we use a different solution algorithm. The authors
in~\cite{Gomes00} suggest adding randomization to the algorithm and
using a rapid restart policy. The restart policy selects a new random
seed for the algorithm and restarts it if the algorithm is not making
sufficient progress with the current seed. 

Finally, we demonstrated the effectiveness of our approach in an
environment where the adversaries had a noncooperative intelligence
that was unknown. We found that the simple model used for the
adversaries in the solver could be mitigated by a multi-level
replanning architecture. In this architecture, there are two levels:
low level task completion and high level task assignment. When
replanning does not occur at either level, the solver fails because it
generates a plan that becomes obsolete as the adversaries use their
intelligence.  When replanning occurs at the task completion level, an
assignment is generated once by the solver. As the adversaries use
their intelligence, the task completion component is run periodically
for each vehicle, generating a new trajectory to complete the tasks in
the vehicle's assignment.  This was somewhat effective at handling
the unknown intelligence.
When replanning occurs at both levels, the
task assignment component is run periodically  in addition to the task
completion component.  We found this replanning architecture effective
at retasking in the dynamically changing environment. It is
advantageous to replan frequently, on average, but there are instances
where replanning frequently is not advantageous. In these cases, the
vehicles are retasked so frequently that their productivity is
reduced.  Therefore, it may be desirable to place a penalty on
changing each vehicle's current task. 

In general, we feel the multi-level replanning approach is a natural
way to handle multi-vehicle cooperative control problems. There are
many different directions for further research, including the addition
of a high level learning module to generate better models of the
adversaries through experience~\cite{Stone00}.

\appendix

\section{Vehicle Dynamics}
\label{sec:vehicleDynamics}
The wheeled robots of Cornell's RoboCup
Team~\cite{Stone01} are the defenders in the RoboFlag problems we
consider in this paper.  We state their governing equations and
simplify them by restricting the allowable control
inputs~\cite{Nagy04}.  The result is a linear set of governing
equations coupled by a nonlinear constraint on the control input.
This procedure allows real-time calculation of many near-optimal
trajectories and has been successfully used by Cornell's RoboCup
team~\cite{Stone01,Nagy04}.  

Each vehicle has a three-motor omni-directional drive
which allows it to  move along any direction irrespective of its
orientation.  This allows for superior maneuverability compared to
traditional nonholonomic (car-like) vehicles.  The nondimensional
governing equations for each vehicle are given by
\begin{equation}
  \left[ \begin{array}{c}
    \ddot{x}(t) \\
    \ddot{y}(t) \\
    \ddot{\theta}(t)
  \end{array} \right] +
  \left[ \begin{array}{c}
    \dot{x}(t) \\
    \dot{y}(t) \\
    \frac{2mL^2}{J}\dot{\theta}(t)
  \end{array} \right] =
  \mathbf{u}(\theta(t),t),
\end{equation}
where $(x(t),y(t))$ are the coordinates of the robot on the playing
field,
$\theta(t)$ is the orientation of the robot, and
$\mathbf{u}(\theta(t),t) = \mathbf{P}(\theta(t)) \mathbf{U}(t)$ can be
thought of as a $\theta(t)$-dependent control input, where
\begin{equation}
  \mathbf{P}(\theta) =
  \left[ \begin{array}{ccc}
    -\sin(\theta)&-\sin(\frac{\pi}{3}-\theta)&\sin(\frac{\pi}{3}+\theta)
\\
    \cos(\theta)&-\cos(\frac{\pi}{3}-\theta)&-\cos(\frac{\pi}{3}+\theta)
\\
    1 & 1 & 1
  \end{array} \right],
\end{equation}
and
\begin{equation}
  \mathbf{U}(t) =
  \left[ \begin{array}{c}
  U_1(t) \\
  U_2(t) \\
  U_3(t)
  \end{array} \right].
\end{equation}
In the equations above, $m$ is the mass of the vehicle, $J$ is the
vehicle's moment of inertia, $L$ is the distance from the drive to the
center of mass, and $U_i(t)$ is the voltage applied to motor $i$.

By restricting the admissible control inputs we simplify the governing
equations in a way that allows near-optimal performance.  The set of
admissible voltages $\mathcal{U}$ is given by the unit cube and the
set of admissible control inputs is given by $P(\theta) \mathcal{U}$.
The restriction involves replacing the set $P(\theta)\mathcal{U}$ with
the maximal $\theta$-independent set found by taking the intersection
of all possible sets of admissible controls.  This set is
characterized by the inequalities
\begin{equation}
  \label{constraint1}
  u_x(t)^2 + u_y(t)^2 \leq \left( \frac{3-|u_\theta(t)|}{2} \right)^2
\end{equation}
and
\begin{equation}
  \label{constraint2}
  |u_\theta(t)| \leq 3,
\end{equation}
where the $\theta$-independent control is given by $(u_x(t), u_y(t),
u_z(t))$.  The equations of motion become
\begin{equation}
  \left[ \begin{array}{c}
    \ddot{x}(t) \\
    \ddot{y}(t) \\
    \ddot{\theta}(t)
  \end{array} \right] +
  \left[ \begin{array}{c}
    \dot{x}(t) \\
    \dot{y}(t) \\
    \frac{2mL^2}{J}\dot{\theta}(t)
  \end{array} \right] =
  \left[ \begin{array}{c}
    u_x(t) \\
    u_y(t) \\
    u_\theta(t)
  \end{array} \right],
\end{equation}
subject to constraints~(\ref{constraint1}) and~(\ref{constraint2}),
which couple the degrees of freedom.  To decouple the $\theta$
dynamics we set $|u_\theta(t)| \leq 1$. Then
constraint~(\ref{constraint1}) becomes
\begin{equation}
  \label{nl_u_constraint}
  u_x(t)^2 + u_y(t)^2 \leq 1.
\end{equation}
Now the equations of motion for the translational dynamics of the
vehicle are given by
\begin{eqnarray}
&\ddot{x}(t) + \dot{x}(t) = u_x(t)\nonumber\\%
&\ddot{y}(t) + \dot{y}(t) = u_y(t),
\label{eom}
\end{eqnarray}
subject to constraint~(\ref{nl_u_constraint}).
In state space form we have
\begin{equation}
  \dot{\mathbf{x}}(t) = \mathbf{A}_c \mathbf{x}(t) + \mathbf{B}_c
  \mathbf{u}(t),
\end{equation}
where $\mathbf{x} = (x,y,\dot{x},\dot{y})$ is the state and
$\mathbf{u} = (u_x,u_y)$ is the control input.


\end{document}